\documentclass{article}

\usepackage{xcolor}

\usepackage[final]{corl_2026} 

\usepackage{amsmath}
\usepackage{amsfonts}
\usepackage{amssymb}
\usepackage{booktabs}
\usepackage{float}
\usepackage{multirow}
\usepackage{tabularx}
\usepackage{caption}
\usepackage{tikz}
\usepackage{threeparttable}
\usepackage{placeins}
\usepackage{dashrule}
\usepackage{graphicx}
\usepackage{wrapfig}
\usepackage{enumitem}
\usepackage{afterpage}

\newcommand{\circled}[1]{\tikz[baseline=(char.base)]{
    \node[shape=circle,draw,inner sep=1pt] (char) {#1};}}
    
\definecolor{fig_red}{RGB}{224, 102, 102}
\definecolor{fig_blue}{RGB}{66,133,244}
\definecolor{realpurple}{RGB}{153, 0, 255}
\definecolor{truemagenta}{RGB}{255, 0, 255}
\definecolor{electriccyan}{RGB}{0, 255, 255}

\usepackage{todonotes}

\newcommand{\notesup}[1]{#1}  

\title{WARPED: Wrist-Aligned Rendering for Robot Policy Learning from
       Egocentric Human Demonstrations}

\author{
  Harry Freeman\thanks{Corresponding author.}
  \qquad Chung Hee Kim
  \qquad Jenny Wang
  \qquad George Kantor
  \\[0.7em]
  Carnegie Mellon University
  \\[0.7em]
  \href{https://kantor-lab.github.io/WARPED/}{\texttt{https://kantor-lab.github.io/WARPED/}}
}

\begin{document}
\maketitle


\begin{abstract}
    Recent advancements in learning from human demonstration have shown promising
results in addressing the scalability and high cost of data collection required
to train robust visuomotor policies. However, existing approaches are often
constrained by a reliance on multiview camera setups, depth sensors, or custom
hardware, and are typically limited to policy execution from third-person or
egocentric cameras. In this paper, we present WARPED, a framework designed to
synthesize realistic wrist-view observations and actions from human
demonstration videos to facilitate the training of visuomotor policies 
using
only monocular RGB data. With data collected from an egocentric RGB camera, our
system leverages vision foundation models to initialize the interactive scene.
A hand-object interaction pipeline is then employed to track the hand and
manipulated object and retarget the trajectories to a robotic end-effector.
Lastly, photorealistic wrist-view observations are synthesized via Gaussian
Splatting to directly train a robotic policy. We demonstrate that WARPED
achieves success rates \notesup{comparable to teleoperation on four of five} rigid-object tabletop tasks and one articulated-object task, while requiring 
\notesup{4-9}$\times$
less data collection time.
\end{abstract}


\section{Introduction}
\vspace{-5pt}
\begin{wrapfigure}{r}{0.6\textwidth}
    \vspace{-3.25\baselineskip}
    \includegraphics[width=\linewidth]{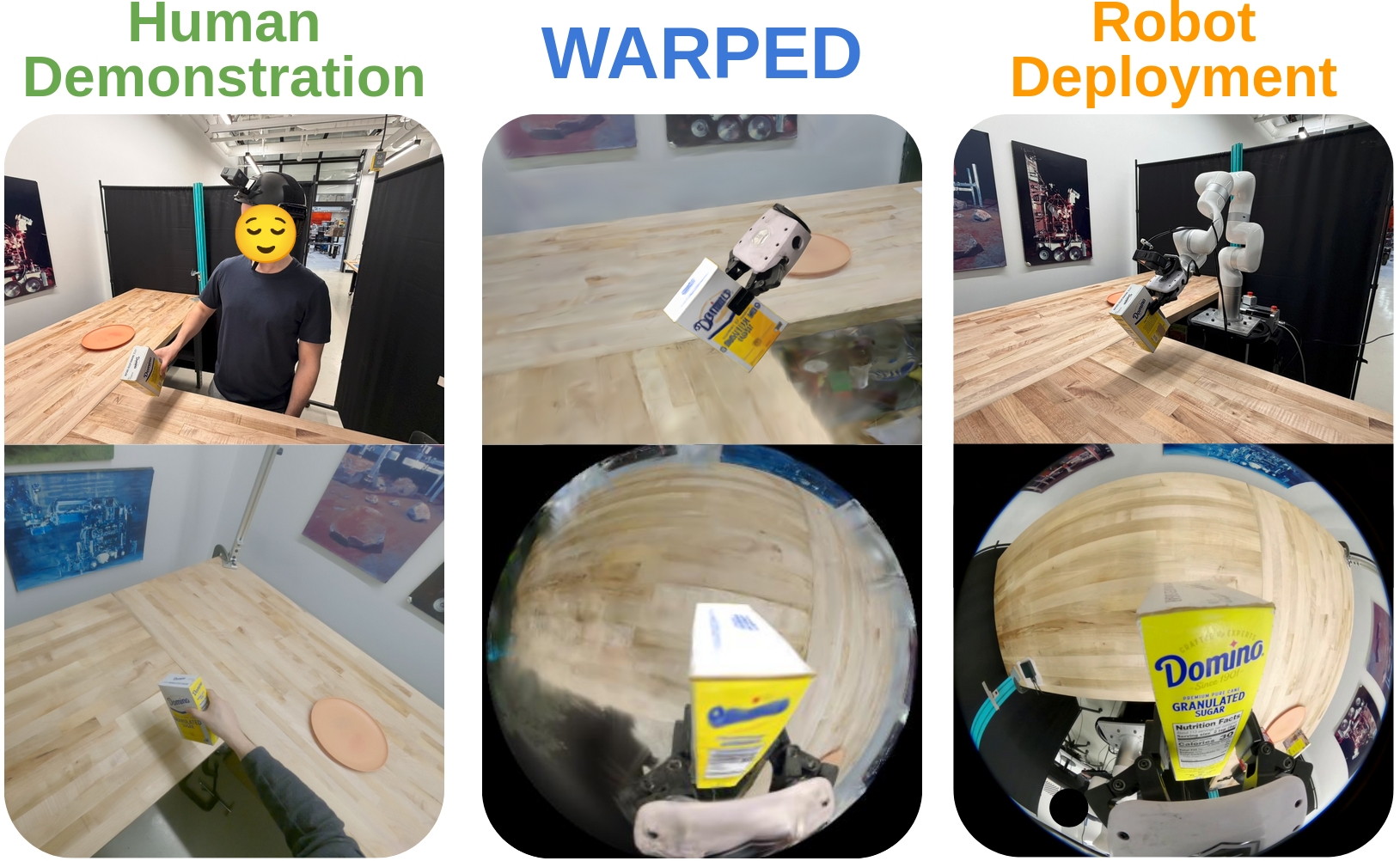}
    \caption{WARPED: A framework that warps egocentric human demonstrations into wrist-camera observations and trajectories for training robot policies. The top row third-person views are for visualization only and are not used as part of the pipeline.}
    \vspace{-6pt}
    \label{fig:front_page}
\end{wrapfigure}

Imitation learning~\cite{motion_tracks_il_0, motion_tracks_il_1, ditto, goyal2024rvt, 3d_diffuser_actor, Ze2024DP3, jia2024lift3dfoundationpolicylifting} has emerged as a popular approach to train robotic visuomotor policies to perform a variety of manipulation tasks. These tasks range from simple pick and place, pushing, and insertion~\cite{jang2021bc, yonemaru2025learningpushgroupgrasp, vecerik2023robotaptrackingarbitrarypoints}, to more complex long-horizon tasks such as folding laundry, tool use, and washing dishes~\cite{pertsch2025fastefficientactiontokenization, chen2025tool, chi2023diffusionpolicy}. However, the performance of these policies is highly dependent on the availability and quality of the demonstration data. Methods often utilize existing large-scale datasets from teleoperated robot demonstrations~\cite{open_x, fang2024rh20t} or internet videos~\cite{epik_kitchens, ego4d}, which are expensive and difficult to collect when scaling to new tasks and environments. 
Alternatively, methods can rely on collecting new teleoperated robot data~\cite{iyer2024openteachversatileteleoperation, shafiullah2023bringingrobotshome, bunnyvisionpro, anyteleop}, which is slow, time-consuming, and labor-intensive to acquire.

Recent works on learning directly from a small number of human demonstrations without teleoperation or large-scale datasets have been proposed to address these limitations and improve the scalability and adaptability of training visuomotor policies. Because humans manipulate objects quickly and naturally, demonstrations can be collected significantly faster than teleoperation, allowing for rapid adaptation to new tasks. To bridge the observation gap between human demonstrations and robot execution, prior approaches have utilized specialized data collection interfaces~\cite{fastumi, wu2024gellogenerallowcostintuitive, act}, data augmentation strategies~\cite{pan20251001demos, mandlekar2023mimicgendatagenerationscalable, garrett2024skillmimicgenautomateddemonstrationgeneration}, and simplified intermediate representations such as hand-pose trajectories~\cite{motiontrans, papagiannis2025rxretrievalexecutioneveryday}, keypoints~\cite{p3po, point_policy, yang2022learningperiodictaskshuman}, and object-centric 3D representations~\cite{rsrd,yu2025real2render2realscalingrobotdata, orion}.

However, many of these approaches still depend on additional sensing or setups that limit how easily data can be collected for new tasks. Several works require multi-view images~\cite{motion_tracks, point_policy, robotube} or depth sensors~\cite{demodiffusion, okami, phantom} for reconstruction and tracking, rely on custom hardware-based collection interfaces~\cite{umi, wang2024dexcapscalableportablemocap}, or depend on custom-trained generative models to convert human demonstrations into robot-compatible observations~\cite{rwor, gen2act}. As a result, if a user wants to train a policy for a new manipulation task, they need one of these components, which may be difficult to set up and use or not readily available. Additionally, many of these methods are limited to rolling out policies on fixed or egocentric camera viewpoints. Wrist-mounted camera viewpoints are often desirable, as they capture more fine-grained details of the interactive scene~\cite{Acar_2024, chuang2025activevisionneedexploring, givingrobotshand}. Because most human demonstration-based approaches require policies to be executed from camera viewpoints similar to those used for data collection, they cannot readily leverage wrist-camera observations at deployment.

To address these limitations, we present \textbf{W}rist-\textbf{A}ligned \textbf{R}endering for Robot \textbf{P}olicy Learning from \textbf{E}gocentric Human \textbf{D}emonstrations (WARPED), an approach for training wrist-camera visuomotor policies from human demonstrations without requiring multi-view sensing, depth sensors, or custom collection hardware. With only a single head-mounted RGB camera, WARPED takes egocentric videos of human demonstrations and produces robot-aligned observations and actions from a wrist-camera perspective for policy learning. Our system leverages vision foundation models to initialize the scene and a hand-object interaction pipeline to track and retarget human trajectories to a robotic end-effector. We then use Gaussian Splatting to render photorealistic wrist-view observations to train a robotic policy. We demonstrate that WARPED achieves \notesup{success rates comparable to teleoperation on four of five rigid-object tabletop tasks and one articulated object task}, while requiring 
\notesup{4-9~times}
less data collection time. This work presents an initial approach and validation for monocular egocentric-to-wrist conversion for short-horizon tabletop manipulation tasks, with the intention of extending to richer contact and longer-horizon tasks in future work. Our main contributions are:
\begin{itemize}[leftmargin=*,noitemsep,topsep=0pt,parsep=0pt,partopsep=0pt]
    \item A pipeline for generating robot-aligned wrist-view observations and trajectories from egocentric human demonstrations to train visuomotor policies.
    \item A monocular RGB-only approach that integrates vision foundation models and Gaussian Splatting to synthesize photorealistic wrist-view observations without depth sensors or custom hardware.
    \item An evaluation and analysis on five tabletop manipulation tasks and one articulated-object task, \notesup{demonstrating similar teleoperation-level success} with significantly reduced data collection time.
\end{itemize}
\vspace{-6pt}

\section{Related Work}
\textbf{Imitation Learning.} Recent progress in imitation learning has focused on learning visuomotor policies directly from visual and proprioceptive inputs~\cite{black2026pi0visionlanguageactionflowmodel, kim2024openvlaopensourcevisionlanguageactionmodel, octomodelteam2024octoopensourcegeneralistrobot, brohan2023rt2visionlanguageactionmodelstransfer, brohan2023rt1roboticstransformerrealworld}, allowing robots to perform increasingly complex manipulation tasks. However, most approaches rely on large amounts of robot-collected data, making them difficult to scale to new tasks and scenes. Expert demonstrations are often collected through teleoperation, where a human controls a robot and records observations and actions. 
Teleoperation can be performed using a wide range of interfaces, including a keyboard~\cite{spacemouse0, spacemouse1}, game controller~\cite{controller_0}, or 3D SpaceMouse~\cite{robomimic2021, 6343868}, which can be limited in fine-grained manipulation. VR and AR-based controllers~\cite{motiontrans, duan2023ar2d2trainingrobotrobot, chen2024arcapcollectinghighqualityhuman, zhang2018deepimitationlearningcomplex, activeumi} enable more visually guided interaction but introduce additional hardware requirements and user learning overhead. Custom hardware systems~\cite{act, fu2024mobilealohalearningbimanual, umi, wu2024gellogenerallowcostintuitive, dexumi} have also been proposed to facilitate data collection, but require specialized equipment.

\textbf{Learning from Human Video.} To reduce the cost of data collection, prior works have explored learning robot policies directly from human demonstrations. Several approaches leverage large-scale human video datasets to pretrain visual representations~\cite{r3m, vip, liv, vc1}, learn object and scene-level affordances~\cite{bahl2023affordanceshumanvideosversatile, vidbot, zeromimic}, or predict robot actions~\cite{bharadhwaj2023zeroshotrobotmanipulationpassive, kumar_1, shaw2022videodexlearningdexterityinternet, egoscale}. Other methods extract coarse supervision from human videos, such as keypoints~\cite{p3po, point_policy, papagiannis2025rxretrievalexecutioneveryday, motion_tracks}, object trajectories~\cite{ditto, okami, orion, yin2025objectcentric3dmotionfield, yoto, spot, lum2025crossinghumanrobotembodimentgap}, correspondence tracks~\cite{track2act, gen2act}, or motion priors~\cite{demodiffusion, singh2024handobjectinteractionpretrainingvideos}. 
Works have also studied learning from egocentric human video~\cite{EgoMimic, egozero, yu2025egomilearningactivevision, egovla, egobridge}, as the demonstrations naturally capture hand-object interactions from a first-person perspective. However, across these approaches, data collection often depends on curated annotations from large-scale datasets, multi-view and RGB-D camera setups, or specialized sensing hardware, which are not always available
for new tasks.

\textbf{View and Embodiment Synthesis for Robot Learning.} Recent works have explored synthesizing robot-centric wrist-camera views from alternative viewpoints for visuomotor policy learning. WristWorld~\cite{wristworld} extends VGGT~\cite{vggt} to generate temporally consistent wrist-view observations from third-person camera inputs. Imagination at Inference~\cite{imatinf} fine-tunes ZeroNVS~\cite{zeronvs} to synthesize auxiliary wrist-camera images during policy execution. \notesup{These works synthesize wrist views from robot observations rather than converting human demonstrations into robot-aligned training data.} RwoR~\cite{rwor} \notesup{converts human demonstrations into UMI~\cite{umi} observations, but uses a wrist-mounted camera and trains a generative model.} \notesup{Other works use visual editing for cross-embodiment transfer, either between robot embodiments~\cite{mirage_inpaint, chen2024roviaugrobotviewpointaugmentation, lepert2025shadowleveragingsegmentationmasks} or from human videos to robot-compatible observations~\cite{h2r, phantom, lepert2025masquerade}. However, these methods either operate on robot demonstrations or replace the human embodiment within the source camera, and do not address the viewpoint shift from egocentric observations to robot wrist-camera views.} 
\notesup{R2R2R~\cite{yu2025real2render2realscalingrobotdata} similarly generates robot training data from a monocular human demonstration, but centers on object-centric trajectory extraction followed by robot kinematic synthesis.}
\notesup{Generative data augmentation has also been used to synthesize novel viewpoints, scenes, trajectories, or robot executions from limited real demonstrations}~\cite{tian2025viewinvariantpolicylearningzeroshot, xu2025egodemogennovelegocentricdemonstration, real2gen, yang2025noveldemonstrationgenerationgaussian, pan20251001demos, zhang2024diffusionmeetsdaggersupercharging, zhu2025nerfaugdataaugmentationrobotics, yu2023scalingrobotlearningsemantically}.
\notesup{These approaches use neural scene representations, generative models, or simulation to expand training data without collecting corresponding real-world robot trajectories.}
\notesup{Instead, WARPED converts monocular egocentric human demonstrations into wrist-view observations and retargeted robot actions for policy learning, without requiring depth sensing or multiple cameras.}
\section{Method}

\begin{figure}[t]
    \centering
    \includegraphics[width=1.0\textwidth]{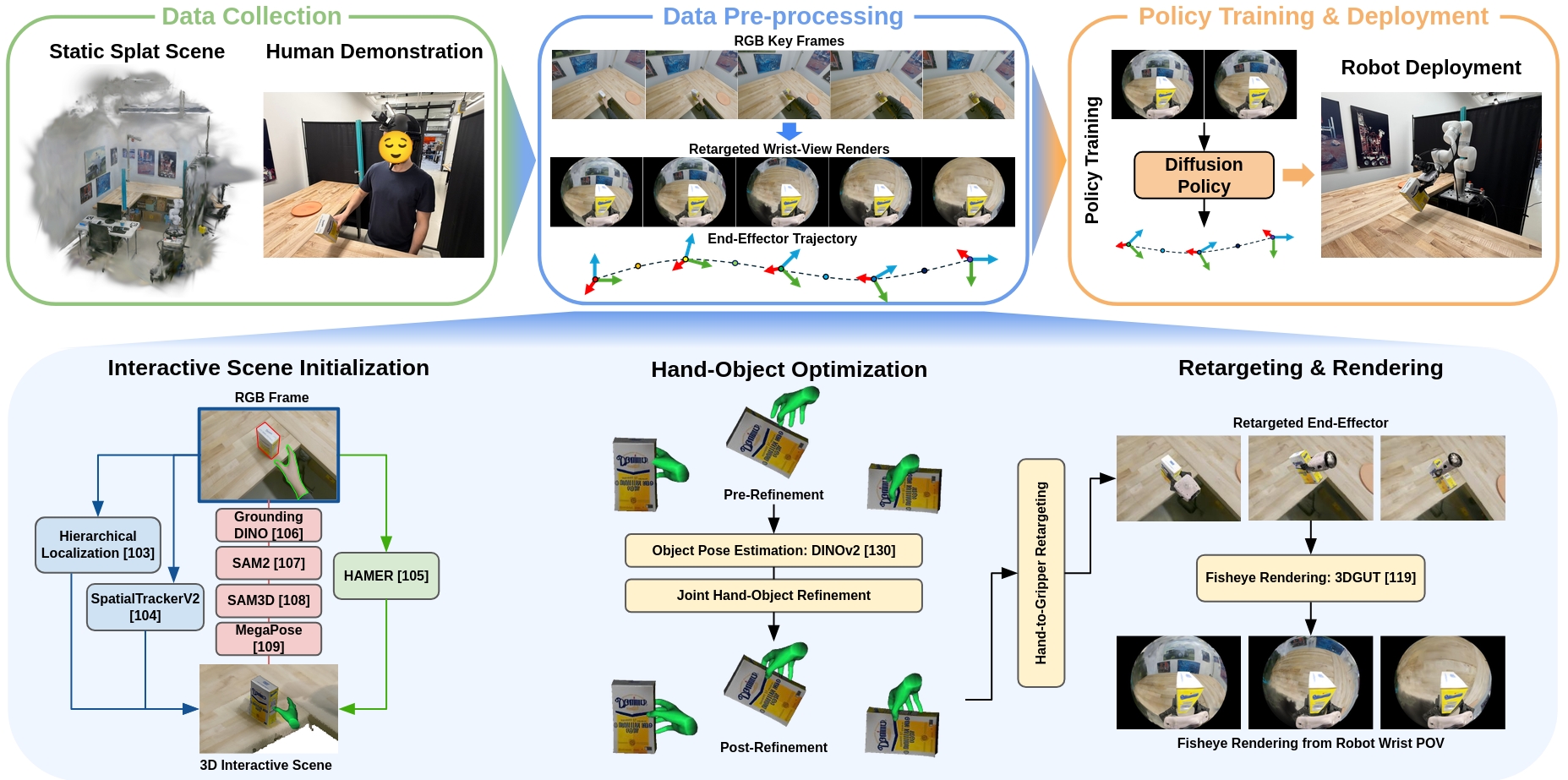}
    \caption{Overview of WARPED. Images of the scene are used to build a Gaussian Splat representation. The user then performs a tabletop manipulation task, recording an egocentric demonstration from a helmet-mounted GoPro. The demonstration is aligned with the Gaussian Splat, and the object pose is localized using vision foundation models. The object pose is then tracked via hand-object optimization. The resulting trajectories are retargeted to a robot gripper, and wrist-view renderings are synthesized and used to train a diffusion policy.}
    \vspace{-8pt}\label{fig:full_pipeline}
\end{figure}

WARPED transforms
RGB egocentric human demonstrations into wrist-camera robotic observation-action datasets to train visuomotor policies for tabletop manipulation tasks. An overview of our pipeline is shown in  Fig.~\ref{fig:full_pipeline}. The pipeline consists of five stages: data collection, interactive scene initialization, hand-object optimization, wrist-view retargeting and rendering, and policy training and deployment. The system is designed to be portable and easy to use, enabling demonstration collection using only a single monocular RGB camera 
without specialized hardware or sensing.

\subsection{Assumptions}
We primarily consider rigid-object tabletop manipulation tasks with a gripper end-effector and without significant scene changes beyond the hand-object interaction. The manipulated object is assumed to remain partially visible, avoiding full occlusions that would prevent tracking. We assume limited in-hand object motion and demonstrations without repeated re-grasping or abrupt shifts in hand-object contact points. We also evaluate on an articulated object whose base remains stationary during demonstration and whose manipulated part is limited to one dominant axis of rotation.

\subsection{Data Collection}
\label{subsec:data_collection}
The user first records a short RGB scan of the workspace without the manipulated objects, analogous to the environment scan used by UMI~\cite{umi}. The scan typically takes less than one minute. Structure-from-Motion (SfM) with LightGlue~\cite{lightglue} feature matching is then used to estimate camera poses and build a sparse 3D reconstruction of the environment. The resulting camera poses are used to initialize a 3D Gaussian Splat of the scene and to localize subsequent demonstrations.
The user then records task demonstrations in the workspace wearing a head-mounted egocentric camera. In our experiments, we use a GoPro Hero9 with a standard linear lens, modeled as a pinhole camera, although other monocular RGB cameras and mounting configurations could be used.

\subsection{Interactive Scene Initialization}
\label{subsec:active_scene_init}

\textbf{Interactive Scene Reconstruction.}
\label{subsubsec:interactive_scene_recon}
For each demonstration, frames are localized within the static scene using Hierarchical Localization~\cite{hloc} to estimate per-frame camera poses. SpatialTrackerV2~\cite{spatialtrackerv2} is used to extract temporally consistent monocular depth maps, which are scale-aligned with the SfM reconstruction and Gaussian Splat. Additional details are provided in Appendix~\ref{app:ssa}.

\textbf{Hand Pose Initialization.}
\label{subsubsec:hand_pose_est}
 HAMER~\cite{hamer} is used to obtain initial per-frame hand shape and pose estimates. The hand parameters are refined using a sequence-level optimization that enforces temporal smoothness and monocular depth consistency. A complete description can be found in Appendix~\ref{app:hpi}.

\textbf{Object Pose Initialization.}
\label{subsubsec:object_pose_init}
Given a text description of the manipulated object, Grounding DINO~\cite{grounding_dino} detects the object in the initial frame. SAM2~\cite{sam2} then generates object masks and propagates them through the sequence. For each task, SAM3D~\cite{sam3d_obj} reconstructs an initial object mesh, from which we construct an object Gaussian Splat. Finally, MegaPose~\cite{megapose} uses the mesh and first-frame segmentation to initialize the 6D object pose for each demonstration. The object scale and pre-contact pose are then refined using segmentation and monocular-depth consistency. Full details are provided in Appendix~\ref{app:obj_pose_init}.

\subsection{Hand-Object Optimization}
\label{subsec:ho_opt} 
\begin{figure}[t]
    \centering
    \includegraphics[width=0.86\linewidth]{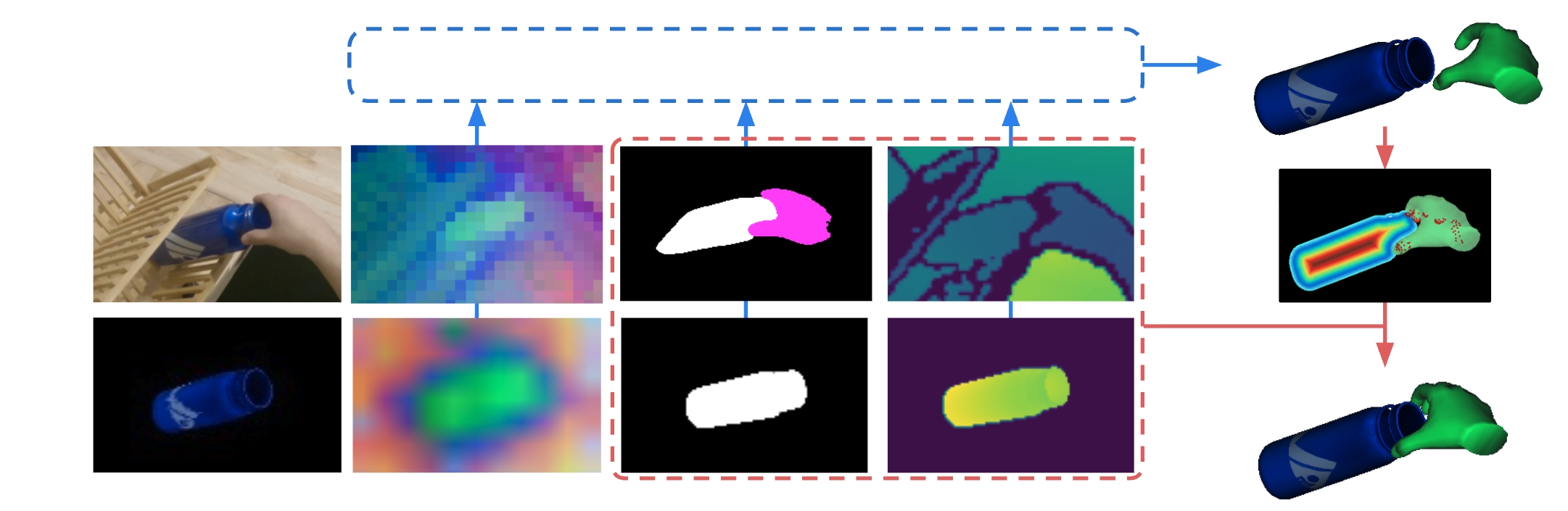}
    \setlength{\unitlength}{1cm}
    \begin{picture}(0,0)
    \put(-8.90, -0.05){\scriptsize \textbf{DINOv2}}
    \put(-6.85, -0.05){\scriptsize \textbf{Mask}}
    \put(-4.80, -0.05){\scriptsize \textbf{Depth}}
    \put(-12.00, 2.05){\rotatebox[origin=c]{90}{\scriptsize \textbf{Demo}}}
    \put(-11.77, 2.05){\rotatebox[origin=c]{90}{\scriptsize \textbf{Frame}}}
    \put(-12.00, 0.82){\rotatebox[origin=c]{90}{\scriptsize \textbf{Rendered}}}
    \put(-11.77, 0.86){\rotatebox[origin=c]{90}{\scriptsize \textbf{Image}}}
    \put(-2.94, 2.01){\rotatebox[origin=c]{90}{\textcolor{fig_red}{\scriptsize \textbf{Interaction}}}}
    \put(-2.71, 2.01){\rotatebox[origin=c]{90}{\textcolor{fig_red}{\scriptsize \textbf{Constraints}}}}
    \put(-2.64, 4.05){\textcolor{fig_blue}{\scriptsize \textbf{\circled{1} Object Pose}}}
    \put(-2.25, 3.78){\textcolor{fig_blue}{\scriptsize \textbf{Estimation}}}
    \put(-0.94, 1.00){\textcolor{fig_red}{\scriptsize \textbf{\circled{2} Hand-Object}}}
    \put(-0.47, 0.70){\textcolor{fig_red}{\scriptsize \textbf{Refinement}}}
    \put(-8.90, 3.39){\textcolor{fig_blue}{\scriptsize $\mathcal{L}_{\text{DINO}}$}}
    \put(-6.75, 3.39){\textcolor{fig_blue}{\scriptsize $\mathcal{L}_{\mathcal{M}}$}}
    \put(-4.60, 3.39){\textcolor{fig_blue}{\scriptsize $\mathcal{L}_{\mathcal{D}}$}}
    \end{picture}
    \caption{Overview of hand-object optimization. The object pose is first estimated using mask, depth, and DINOv2 losses. The hand and object pose are then jointly refined using mask, depth, and interaction constraints.}
    \vspace{-8pt}\label{fig:ho_opt}
\end{figure}
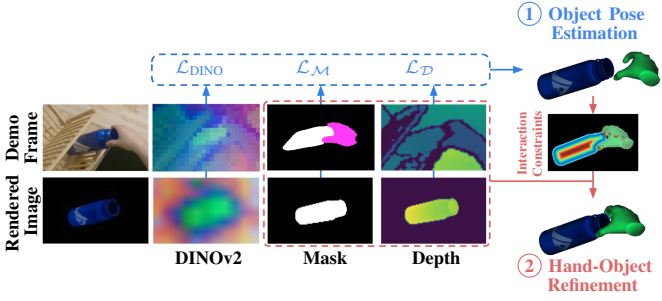

To render object-centric views, we need to track the 3D pose of the object. While prior works rely on depth sensors~\cite{papagiannis2025rxretrievalexecutioneveryday, p3po, demodiffusion}, multi-view cameras~\cite{motion_tracks, point_policy, phantom}, smart glasses~\cite{egozero}, or full 3D object scans~\cite{rsrd, yu2025real2render2realscalingrobotdata}, our system assumes none of these.
As a result, we adopt a hand-object interaction-based approach, leveraging the fact that hand and object motion provide complementary geometric constraints.
Similar to prior works~\cite{hobman, rhov, hold, rhit}, we utilize two stages~(Fig.~\ref{fig:ho_opt}), where the object pose is first estimated independently and then jointly refined together with the hand. 

\textbf{Object Pose Estimation.}
Given the object mesh, object masks, monocular depth maps, and the pre-contact pose from Sec.~\ref{subsec:active_scene_init}, the object pose is estimated sequentially using differentiable rendering. For each frame, a differentiable rasterizer renders the object image, mask, and depth. We optimize each pose using an occlusion-aware mask loss~\cite{phosa} $\mathcal{L}_{\mathcal{M}_{obj}}$, monocular depth loss $\mathcal{L}_{\mathcal{D}_{obj}}$, and a DINOv2 feature loss $\mathcal{L}_{\text{DINO}}$ between the rendered object and masked demonstration frame.
\begin{equation}
\label{eq:obj_pose_opt}
\min_{\mathbf{R}^{obj}_t, \mathbf{t}^{obj}_t}
\;
\lambda_{\mathcal{M}_{obj}} \mathcal{L}_{\mathcal{M}_{obj}}
+
\lambda_{\mathcal{D}_{obj}} \mathcal{L}_{\mathcal{D}_{obj}}
+
\lambda_{\text{DINO}} \mathcal{L}_{\text{DINO}} .
\end{equation}
Each pose is optimized independently using the previous pose as initialization. Contact start and end frames $(t_s, t_e)$ are estimated using object motion thresholds over consecutive frames. Details of the losses and contact estimation are provided in Appendix~\ref{app_subsec:ope}-\ref{app:start_end}.

\textbf{Joint Hand-Object Refinement.}
Because estimating hand and object poses independently can be inaccurate, we jointly optimize both to exploit hand-object interaction constraints. The hand mesh is generated from MANO~\cite{mano} using the current hand parameters, and a differentiable rasterizer renders hand and object masks and depths across all frames.

We apply occlusion-aware mask losses to both the object and hand, $\mathcal{L}_{\mathcal{M}}=\mathcal{L}_{\mathcal{M}_{obj}}+\mathcal{L}_{\mathcal{M}_{hand}}$, and monocular depth losses, $\mathcal{L}_{\mathcal{D}}=\mathcal{L}_{\mathcal{D}_{obj}}+\mathcal{L}_{\mathcal{D}_{hand}}$. To encourage hand-object interaction, a contact loss $\mathcal{L}_{\text{contact}}$ keeps frequently contacted hand vertices~(Fig.~\ref{fig:retarget_and_render}(a)) close to the object surface during manipulation. To discourage physically implausible interpenetration, we use a collision loss $\mathcal{L}_{\text{col}}$ that penalizes hand vertices inside the object volume. A stable grasp loss~\cite{rhit} $\mathcal{L}_{\text{sg}}$ further encourages locally consistent fingertip-object distances during contact. The final joint optimization is performed simultaneously across all frames
\begin{equation}
\label{eq:joint_ho}
\min_{\boldsymbol{\Theta}} \; 
\lambda_{\mathcal{M}} \mathcal{L}_{\mathcal{M}}
+ \lambda_{\mathcal{D}} \mathcal{L}_{\mathcal{D}}
+ \lambda_c \mathcal{L}_{\text{contact}}
+ \lambda_{\text{col}} \mathcal{L}_{\text{col}}
+ \lambda_{\text{sg}} \mathcal{L}_{\text{sg}}
+ \mathcal{L}_{\text{aux}}
\end{equation}
where $\boldsymbol{\Theta}=\{\boldsymbol{\Theta}^{obj}, \boldsymbol{\Theta}^{hand}\}$ denotes the optimized object and hand parameters, and $\mathcal{L}_{\text{aux}}$ represents additional auxiliary scene and hand regularization terms. The object is constrained to remain stationary outside the estimated contact interval $(t_s,t_e)$. Full definitions are provided in Appendix~\ref{app:joint_ho}-\ref{app_subsec:aux_loss}.

\textbf{Articulated-Object Extension.}
For an articulated object with a single dominant axis of rotation, we optimize the joint angle of the manipulated part rather than a full 6D object pose. We first partition the object Gaussian Splat into base and manipulated-part splats using a part-level segmentation of the reconstructed object. Instead of a differentiable rasterizer, the object is rendered using Gaussian Splatting, where splats associated with the object base remain fixed during the demonstration and splats associated with the manipulated part are transformed according to the joint angle.

\subsection{Retargeting and Rendering}
\label{subsec:retarg_and_red}
Once the hand and object poses have been recovered through joint optimization, the hand-object motion is retargeted into executable robot end-effector trajectories, and the corresponding wrist-camera observations are rendered. First, a sparse set of evenly spaced keyframes is extracted from the hand-object trajectory. 
For keyframes prior to contact ($t < t_s$), the gripper remains open and the end-effector pose is mapped from the human hand using the thumb and index joints, as shown in Fig.~\ref{fig:retarget_and_render}(b) and discussed in Appendix~\ref{app_subsec: h2eem}. Following Pan \textit{et al.}~\cite{pan20251001demos}, we optimize the pre-contact trajectory to prevent unintended gripper-object collisions while remaining close to the original motion.

At the contact start frame ($t = t_s$), the end-effector pose and gripper width are refined to ensure a physically plausible grasp~(Fig.~\ref{fig:retarget_and_render}(c-f)). For both the thumb and index fingertips, 50 contact points are identified based on mesh vertex proximity. Using these contact points, the end-effector position and gripper width are optimized to align the grasp with the hand-held object. Additional details are in Appendix~\ref{app_subsec_f2:pcto}-\ref{app_subsec_f4:cgr}. Although our experiments use a parallel-jaw gripper, this retargeting step could be substituted for other end-effectors, such as dexterous robot hands~\cite{anyteleop, dexpilot}.

During contact ($t_s \leq t \leq t_e$), the poses of the gripper and object are assumed to be relatively fixed. For subsequent frames, the end-effector pose is updated by applying the relative object motion with respect to the contact frame,
ensuring that the end-effector rigidly follows the object motion while preserving the established grasp.
Finally, keyframe poses are interpolated to produce smooth trajectories. Wrist-camera views of the retargeted trajectories are rendered by combining the Gaussian Splats of the scene, object, and end-effector using Nerfstudio's~\cite{nerfstudio, gsplat} 3DGUT~\cite{3dgut} implementation. Example renderings are shown in Fig.~\ref{fig:retarget_and_render}(right).

\begin{figure}[t]
    \centering
    \includegraphics[width=\linewidth]{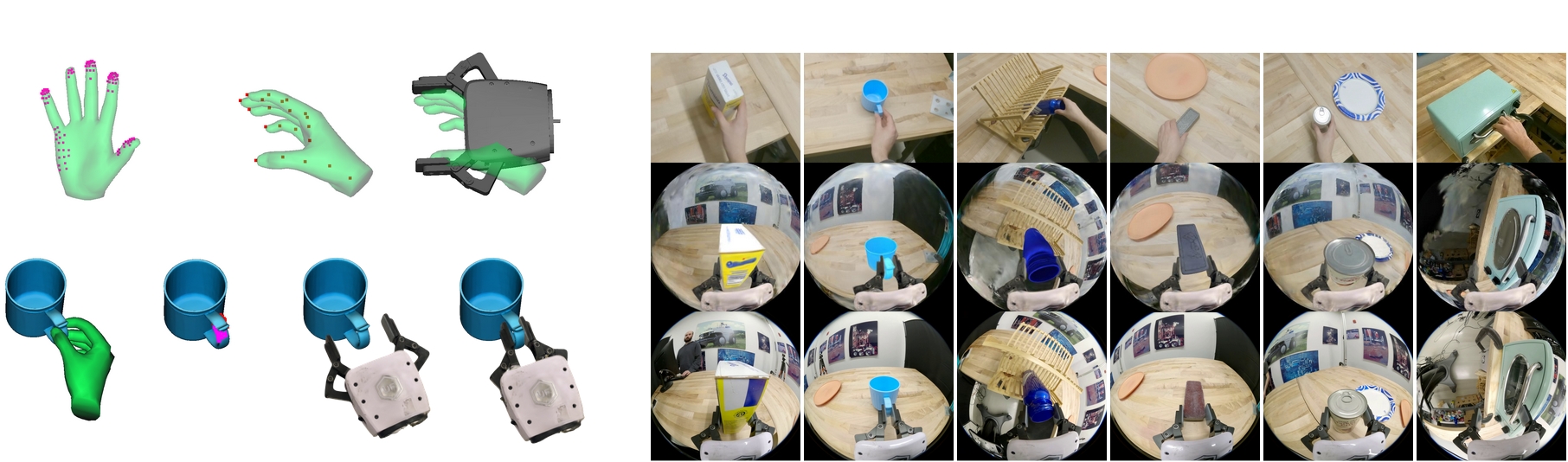}
    \setlength{\unitlength}{1cm}
    \begin{picture}(0,0)
        \put(-6.4, 2.35){\footnotesize (a)}
        \put(-3.6, 2.35){\footnotesize (b)}
        \put(-6.6, 0.25){\footnotesize (c)}
        \put(-5.3, 0.25){\footnotesize (d)}
        \put(-4.1, 0.25){\footnotesize (e)}
        \put(-2.7, 0.25){\footnotesize (f)}
        \put(-3.65, 3.25){\textbf{$\rightarrow$}}
        \put(-6.15, 1.7){\textbf{$\rightarrow$}}
        \put(-4.85, 1.7){\textbf{$\rightarrow$}}
        \put(-3.45, 1.7){\textbf{$\rightarrow$}}

        \put(-1.03, 4.43){\footnotesize \textbf{Rotate}}
        \put(-0.85, 4.1){\footnotesize \textbf{Box}}
        \put(0.45, 4.43){\footnotesize \textbf{Pour}}
        \put(0.50, 4.1){\footnotesize \textbf{Mug}}
        \put(1.75, 4.43){\footnotesize \textbf{Bottle}}
        \put(1.52, 4.1){\footnotesize \textbf{from Rack}}
        \put(3.15, 4.43){\footnotesize \textbf{Wipe}}
        \put(3.15, 4.1){\footnotesize \textbf{Brush}}
        \put(4.55, 4.43){\footnotesize \textbf{Can}}
        \put(4.35, 4.1){\footnotesize \textbf{on Plate}}
        \put(5.90, 4.43){\footnotesize \textbf{Open}}
        \put(5.60, 4.1){\footnotesize \textbf{Microwave}}

        \put(-1.45, 3.4){\rotatebox[origin=c]{90}{\scriptsize \textbf{Demo}}}
        \put(-1.45, 2.25){\rotatebox[origin=c]{90}{\scriptsize \textbf{Rendered}}}
        \put(-1.45, 0.9){\rotatebox[origin=c]{90}{\scriptsize \textbf{Real}}}
    \end{picture}
    \caption{
    Left: (a) Frequently contacted hand vertices. (b) Pose mapping from hand to end-effector using thumb and index joints. At the contact start frame (c), the closest 50 object points to the thumb and index fingertips (d) are used to refine the initial gripper pose and width (e) into a configuration that establishes contact with the points (f). Right: Example demonstration frames, wrist-view renders, and real rollout images at initial contact.
    }
    \vspace{-8pt}\label{fig:retarget_and_render}
\end{figure}
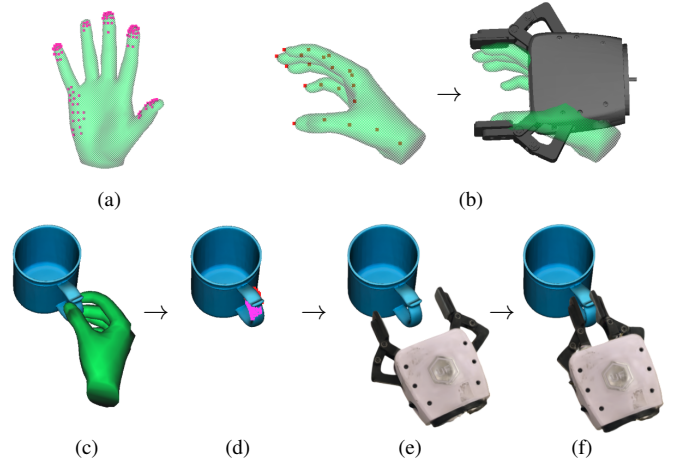

\subsection{Policy Training}
\label{policy_training}

\begin{wrapfigure}{r}{0.65\textwidth}
    \vspace{-3.25\baselineskip}
    \centering
    \hspace{0.04\linewidth}%
    \includegraphics[width=0.96\linewidth]{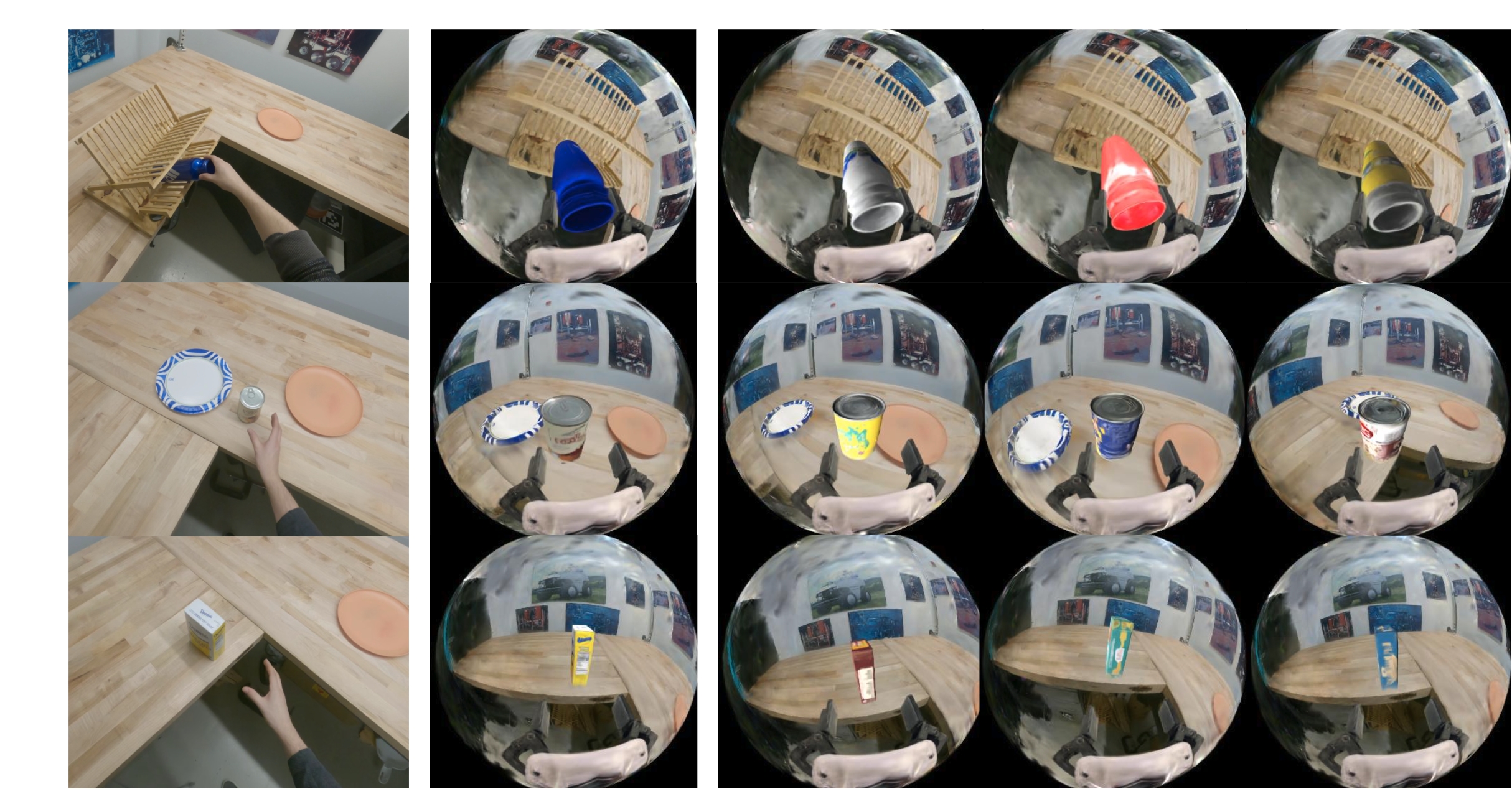}
    \setlength{\unitlength}{1cm}
    \begin{picture}(0,0)
        \put(-3.75, 4.85){\footnotesize \textbf{Demonstration}}
        \put(-1.45, 4.85){\footnotesize \textbf{Original}}
        \put(0.9, 4.85){\footnotesize \textbf{Augmented Render}}

        \put(-4.35, 3.80){\rotatebox[origin=c]{90}{\scriptsize \textbf{Object}}}
        \put(-4.05, 3.80){\rotatebox[origin=c]{90}{\scriptsize \textbf{Appearance}}}

        \put(-4.35, 2.35){\rotatebox[origin=c]{90}{\scriptsize \textbf{Object}}}
        \put(-4.05, 2.35){\rotatebox[origin=c]{90}{\scriptsize \textbf{Pose}}}

        \put(-4.35, 0.95){\rotatebox[origin=c]{90}{\scriptsize \textbf{Gripper}}}
        \put(-4.05, 0.95){\rotatebox[origin=c]{90}{\scriptsize \textbf{Pose}}}
    \end{picture}
    \vspace{-12pt}
    \caption{Example demonstration frames and wrist-view renders illustrating object appearance, object pose, and gripper pose augmentations.}
    \vspace{-6pt}
    \label{fig:augmentation}
\end{wrapfigure}

We train a diffusion policy~\cite{chi2023diffusionpolicy, umi} conditioned on the wrist-view observations and robot proprioceptive inputs to generate relative pose and gripper action chunks. Similar to prior work~\cite{nerf2real, splatsim}, we add Gaussian noise to the input images for both training and inference to mitigate the sim-to-real gap between rendered wrist-view observations and real robot images. To increase dataset diversity, we augment the demonstrations as shown in Fig.~\ref{fig:augmentation}. Augmentations include retexturing object meshes, translating the initial object pose and interpolating the corresponding object and end-effector trajectories, randomizing the initial gripper pose and interpolating it back into the nominal trajectory, randomly scaling the scene, and perturbing both the wrist-camera intrinsics and extrinsics. Additional details are provided in Appendix~\ref{app:augmentation}.
\newpage
\section{Experiments and Results}
\subsection{Experimental Setup}

\begin{wrapfigure}{r}{0.46\textwidth}
    \vspace{-4.5\baselineskip}
    \centering
    \includegraphics[width=0.85\linewidth]{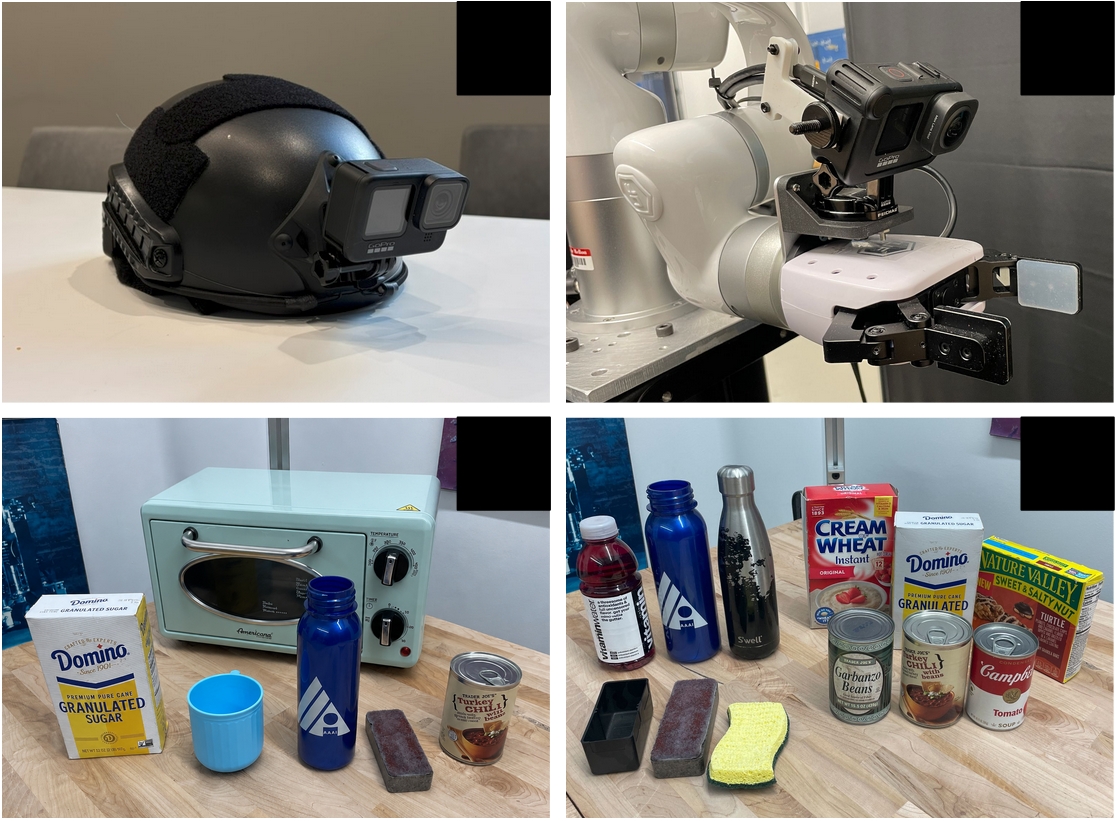}
    \setlength{\unitlength}{1cm}
    \begin{picture}(0,0)
        \put(-3.35, 3.725){\footnotesize\textcolor{white}{(a)}}
        \put(-0.6, 3.725){\footnotesize\textcolor{white}{(b)}}
        \put(-3.35, 1.675){\footnotesize\textcolor{white}{(c)}}
        \put(-0.6, 1.675){\footnotesize\textcolor{white}{(d)}}
    \end{picture}
    \vspace{-2pt}\caption{Experimental setup: (a) helmet-mounted demonstration camera, (b) wrist-mounted rollout camera, (c) task objects, and (d) novel objects.}
    \vspace{-10pt}
    \label{fig:exp_setup}
\end{wrapfigure}

We evaluate WARPED on five rigid-object tabletop manipulation tasks and one articulated-object task using a UFactory xArm7 with an xArm G1 Gripper. For human demonstrations, a GoPro Hero9 is attached to a helmet~(Fig.~\ref{fig:exp_setup}(a)) worn by the user. For policy rollout, a GoPro Hero9 with Max Lens Mod 1.0 is mounted above the gripper to capture wrist-view observations~(Fig.~\ref{fig:exp_setup}(b)). For each task, we collect 30 demonstrations within a fixed workspace using a single object instance, \notesup{and separately capture a single RGB image of the manipulated object for SAM3D mesh reconstruction.} The demonstrations are then processed through WARPED and augmented 10 times~(Sec.~\ref{policy_training}), and a diffusion policy is trained using four NVIDIA Tesla V100s. 
Success rates are evaluated over 20 trials. Further details are provided in Appendix~\ref{app:train_eval}.

\begin{figure}[!h]
    \centering
\includegraphics[width=\linewidth]{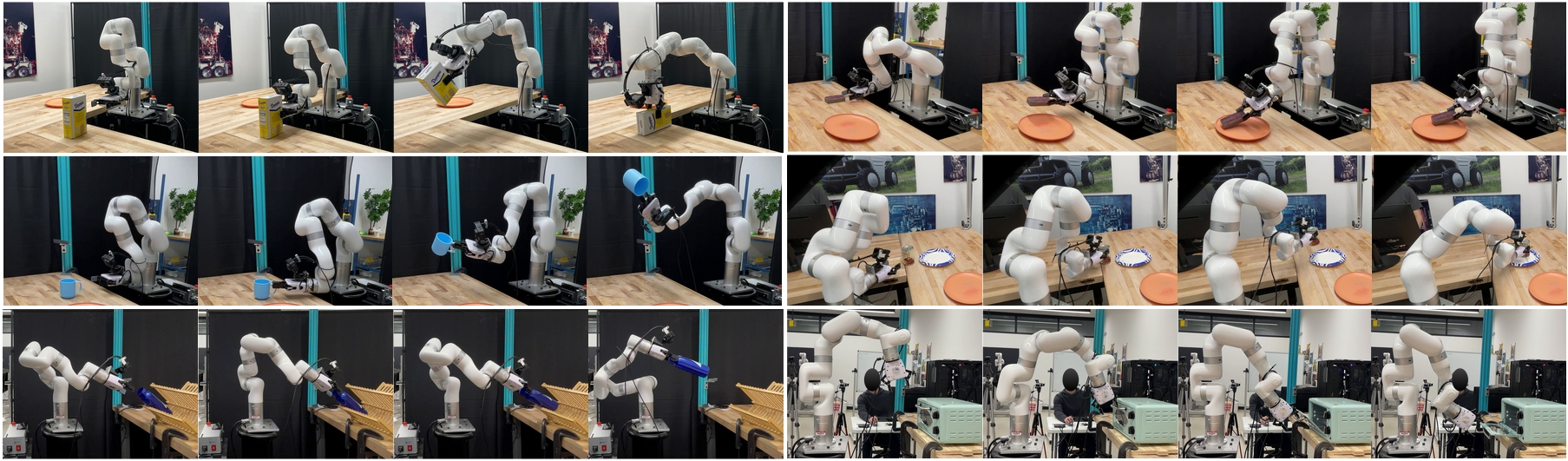}
    \caption{Real-world rollouts for the evaluated tasks.}
    \label{fig:task_rollouts}
    \vspace{-0pt}
\end{figure}

The evaluated tasks are shown in Fig.~\ref{fig:task_rollouts}. Object placement varies across trials. \textbf{Rotate Box:} The robot must rotate a box 90 degrees onto a new face. \textbf{Pour Mug:} The robot must grasp a mug by its handle and execute a pouring motion. \textbf{Bottle from Dish Rack:} The robot must pick up a bottle from the middle shelf of a dish rack. \textbf{Wipe Plate with Brush:} The robot must pick up a brush and wipe a plate. \textbf{Can on Plate:} The robot must pick up a can and place it onto an additionally localized plate. \textbf{Open Microwave:} The robot must pull the microwave door open by its handle.

We compare WARPED against policies trained from teleoperated demonstrations collected with a Meta Quest 2 VR headset and controller. We additionally evaluate against the Alter baseline presented in RwoR~\cite{rwor}. 
\notesup{The teleoperation and Alter baselines also use 30 demonstrations, but do not use the synthetic demonstration augmentations from Sec.~\ref{policy_training}. Generating additional trajectories and observations from the original demonstrations is an advantage of WARPED’s synthetic data pipeline.}
All methods are evaluated with the same single wrist-camera observation modality. We note that RwoR is a relevant baseline, but no public code or trained models were available at the time of testing. 

\subsection{\notesup{Results}}
\label{subsec:results}

 \textbf{Policy Performance.}
We evaluate the performance of our framework where training and evaluation are performed in a single scene using the same objects. The results are shown in Table~\ref{table:policy_capability}. Notably, WARPED achieves \notesup{success rates comparable to teleoperation} on five out of six tasks, despite relying solely on monocular egocentric human demonstrations. However, performance is lower on Wipe Brush. This is because the brush is small and level to the table~(Fig.~\ref{fig:task_rollouts}), making pose estimation and retargeting more difficult. The Alter baseline performs poorly across all tasks, suggesting that wrist-view rendering without accurate hand-object geometry is insufficient for effective policy learning. A two-task comparison with UMI on a UR5 setup is provided in Appendix~\ref{app:umi_comp}.

\begin{table*}[t]
\caption{Success rates comparing teleoperation, the Alter baseline, and WARPED. \notesup{Baselines are unaugmented.}}
\centering
\setlength{\tabcolsep}{9pt}
\renewcommand{\arraystretch}{1.12}
\resizebox{\textwidth}{!}{%
\begin{tabular}{@{}lcccccc@{}} 
\hline\hline
\textbf{Method} & \textbf{Rotate Box} & \textbf{Pour Mug} & \textbf{Bottle from Rack} & \textbf{Wipe Brush} & \textbf{Can on Plate} & \textbf{Open Microwave} \\ 
\hline
Teleoperation & 16/20 & \textbf{19/20} & 16/20 & \textbf{15/20} & \textbf{19/20} & \textbf{18/20} \\
Alter (Heng \textit{et al.}) & 7/20 & 3/20 & 0/20 & 0/20 & 8/20 & 0/20 \\
WARPED & \textbf{20/20} & 18/20 & \textbf{17/20} & 11/20 & 17/20 & 17/20 \\
\hline
\end{tabular}%
}
\vspace{-2pt}
\label{table:policy_capability}
\end{table*}

\begin{wraptable}{r}{0.6\linewidth}
\vspace{-14pt}
\centering
\tiny
\setlength{\tabcolsep}{3pt}
\renewcommand{\arraystretch}{1.05}
\caption{Success rates on novel object instances.}
\vspace{-5pt}
\label{table:novel_objects}
\hspace*{-0.05\textwidth}%
\resizebox{0.49\textwidth}{!}{%
\begin{tabular}{@{}lccccc@{}}
\hline\hline
\textbf{Method} & \multirow{2}{*}{{\shortstack[c]{\textbf{Novel}\\\textbf{Obj}}}}  & \multirow{2}{*}{{\shortstack[c]{\textbf{Rotate}\\\textbf{Box}}}} & \multirow{2}{*}{{\shortstack[c]{\textbf{Bottle}\\\textbf{from Rack}}}} & \multirow{2}{*}{{\shortstack[c]{\textbf{Wipe}\\\textbf{Brush}}}} & \multirow{2}{*}{{\shortstack[c]{\textbf{Can on}\\\textbf{Plate}}}} \\
&&&&& \\
\hline
Teleoperation & Obj 1 & 8/10 & 4/10  & \textbf{7/10} & 9/10 \\
              & Obj 2 & 2/10 & 2/10 & 4/10 & \textbf{9/10} \\
\hline
WARPED & Obj 1 & \textbf{10/10} & \textbf{8/10} & \textbf{7/10} & \textbf{10/10} \\
        & Obj 2 & \textbf{8/10} & \textbf{5/10} & 2/10 & \textbf{9/10} \\
\hline

\end{tabular}%
}
\end{wraptable}

\textbf{Novel Objects.}
We assess WARPED's ability to generalize to novel rigid objects unseen in training~(Fig.~\ref{fig:exp_setup}(d)). The results are shown in Table~\ref{table:novel_objects}. WARPED outperforms teleoperation on all  tasks except Wipe~Brush.

\begin{wraptable}{r}{0.6\linewidth}
\centering
\tiny
\setlength{\tabcolsep}{2pt}
\renewcommand{\arraystretch}{1.10}
\vspace{-15pt}
\caption{Demonstration collection time (MM:SS).}
\vspace{-6pt}
\label{table:data_collection_time}
\resizebox{0.56\textwidth}{!}{%
\begin{tabular}{@{}lcccccc@{}}
\hline\hline
\textbf{Method} &
\multirow{2}{*}{{\shortstack[c]{\textbf{Rotate}\\\textbf{Box}}}} &
\multirow{2}{*}{{\shortstack[c]{\textbf{Pour}\\\textbf{Mug}}}} &
\multirow{2}{*}{{\shortstack[c]{\textbf{Bottle}\\\textbf{from Rack}}}} &
\multirow{2}{*}{{\shortstack[c]{\textbf{Wipe}\\\textbf{Brush}}}} &
\multirow{2}{*}{{\shortstack[c]{\textbf{Can on}\\\textbf{Plate}}}} &
\multirow{2}{*}{{\shortstack[c]{\textbf{Open}\\\textbf{Microwave}}}} \\
&&&&&& \\
\hline
Teleoperation & 22:51 & 24:59 & 31:49 & 30:16 & 15:20 & 35:17 \\
\hline
WARPED & \textbf{3:37} & \textbf{3:18} & \textbf{3:27} & \textbf{5:19} & \textbf{3:41} & \textbf{4:12} \\
\hline
\end{tabular}%
}
\vspace{-5pt}
\end{wraptable}

\textbf{Data Collection Efficiency.}
We evaluate the data collection efficiency of WARPED and compare it to the time required to collect teleoperated data. We report the total time needed
to collect all demonstrations in Table~\ref{table:data_collection_time}. 
For WARPED, this includes the initial scan of the scene. For teleoperation, we report the total time, including environment resets and any additional overhead incurred during operation.
Data 
collection with WARPED is approximately 
\notesup{4-9$\times$}
faster than teleoperation across all tasks. \notesup{This comparison reflects active human data collection time; offline processing requires approximately seven minutes per demonstration and does not require human supervision (Appendix~\ref{app:processing_times})}. The reduced human collection time highlights 
one of WARPED's advantages 
for collecting new datasets, 
and suggests potential for more scalable data collection as complexity and dataset sizes increase.

\textbf{Object Tracking Accuracy.}
We evaluate object tracking accuracy during the contact phase
($t_s~\leq~t~\leq~t_e$) using OptiTrack motion capture to obtain
ground-truth object trajectories. Metrics are computed over 10 trials
per task, separate from the policy rollouts reported in
Table~\ref{table:policy_capability}. Bottle from Rack and Open Microwave are excluded because the rack, hand, or door motion frequently occluded the OptiTrack markers, preventing reliable ground-truth trajectories. 

We report absolute
trajectory error (ATE), relative pose error (RPE) in translation and
geodesic distance over a 0.1s horizon, and mean position error (MPE),
defined as the mean Euclidean distance between predicted and
ground-truth object positions after aligning the contact-start frame as
the anchor.
As shown in Table~\ref{table:object_tracking_accuracy}, WARPED outperforms FoundationPose~\cite{wen2024foundationposeunified6dpose} across all reported metrics,
with the exception of rotational RPE on Rotate Box. Averaged across
tasks, WARPED achieves an ATE of $1.49 \pm 0.61$ cm and an MPE of
$1.81 \pm 0.94$ cm, compared to $7.66 \pm 2.32$ cm and
$11.56 \pm 5.36$ cm for FoundationPose. The weaker performance of
FoundationPose is likely due to noise in monocular depth estimates,
inaccuracies in camera intrinsics, and misalignment in the predicted
object meshes. In contrast, WARPED's hand--object optimization uses
interaction constraints to improve object pose tracking. \notesup{Hand-object optimization ablations and} object pose tracking visualizations are provided in Appendix~\ref{app:ho_ablations} and \notesup{Appendix~\ref{app:obj_pose_track_vis}} respectively.

\begin{table*}[t]
\caption{Object tracking accuracy comparing FoundationPose and WARPED.}
\centering
\setlength{\tabcolsep}{5pt}
\renewcommand{\arraystretch}{1.12}
\resizebox{\textwidth}{!}{%
\begin{tabular}{@{}llcccc@{}}
\hline\hline
\textbf{Method} & \textbf{Metric} &
\textbf{Rotate Box} &
\textbf{Pour Mug} &
\textbf{Wipe Brush} &
\textbf{Can on Plate} \\
\hline
\multirow{4}{*}{FoundationPose}
& ATE (cm) & 5.99 $\pm$ 0.96 & 8.97 $\pm$ 3.12 & 9.16 $\pm$ 0.88 & 6.51 $\pm$ 1.70 \\
& RPE Trans. (cm) & 2.90 $\pm$ 0.30 & 2.42 $\pm$ 0.59 & 1.56 $\pm$ 0.47 & 1.76 $\pm$ 0.26 \\
& RPE Rot. (deg) & \textbf{7.79 $\pm$ 1.45} & 8.15 $\pm$ 1.50 & 3.94 $\pm$ 1.01 & 9.24 $\pm$ 1.82 \\
& MPE (cm) & 6.38 $\pm$ 1.50 & 18.81 $\pm$ 4.16 & 12.68 $\pm$ 1.45 & 8.37 $\pm$ 1.35 \\
\hline
\hline
\multirow{4}{*}{WARPED}
& ATE (cm) & \textbf{1.28 $\pm$ 0.28} & \textbf{2.29 $\pm$ 0.64} & \textbf{1.34 $\pm$ 0.31} & \textbf{1.05 $\pm$ 0.08} \\
& RPE Trans. (cm) & \textbf{0.79 $\pm$ 0.19} & \textbf{1.24 $\pm$ 0.40} & \textbf{0.46 $\pm$ 0.13} & \textbf{0.55 $\pm$ 0.11} \\
& RPE Rot. (deg) & 8.01 $\pm$ 2.53 & \textbf{8.03 $\pm$ 2.70} & \textbf{3.05 $\pm$ 1.13} & \textbf{2.79 $\pm$ 0.49} \\
& MPE (cm) & \textbf{1.43 $\pm$ 0.31} & \textbf{2.37 $\pm$ 0.77} & \textbf{2.60 $\pm$ 0.95} & \textbf{0.85 $\pm$ 0.17} \\
\hline
\end{tabular}%
}
\vspace{-8pt}
\label{table:object_tracking_accuracy}
\end{table*}

\paragraph{\notesup{Render-to-Real Fidelity.}}
To quantify the appearance gap between synthesized and real observations, we uniformly sample 10 wrist-camera frames for every trajectory across all tasks, localize each frame in its reconstructed scene, and render the scene from the corresponding camera viewpoint. We mask the manipulated object as our object-pose estimator does not support fisheye images; \notesup{object tracking is evaluated separately in Table~\ref{table:object_tracking_accuracy}.} We report LPIPS~\cite{lpips} and SSIM~\cite{ssim} between rendered and real views, and contextualize these values using \notesup{Gaussian blur ($\sigma_b=1,2$~pixels) and noise ($\sigma_n=5/255,10/255$) applied to the real images as reference perturbations.}

\newpage
\begin{wraptable}{r}{0.43\linewidth}
\vspace{-0pt}
\centering
\scriptsize
\setlength{\tabcolsep}{5pt}
\renewcommand{\arraystretch}{1.08}
\caption{Render-to-real fidelity comparison. \notesup{Manipulated objects are masked.}}
\vspace{-5pt}
\label{table:render_real_fidelity}
\resizebox{\linewidth}{!}{%
\begin{tabular}{@{}lcc@{}}
\hline\hline
\textbf{Comparison} & \textbf{LPIPS $\downarrow$} & \textbf{SSIM $\uparrow$} \\
\hline
Render--real
    & $.098{\pm}.014$ & $.775{\pm}.048$ \\
$\sigma_b{=}1$
    & $.050{\pm}.004$ & $.973{\pm}.003$ \\
$\sigma_b{=}2$
    & $.127{\pm}.009$ & $.904{\pm}.012$ \\
$\sigma_n{=}5/255$
    & $.071{\pm}.012$ & $.810{\pm}.014$ \\
$\sigma_n{=}10/255$
    & $.173{\pm}.020$ & $.564{\pm}.027$ \\
\hline
\end{tabular}%
}
\vspace{-5pt}
\end{wraptable}

Results are shown in Table~\ref{table:render_real_fidelity}. Across the sampled frames, rendered views achieve an LPIPS of $.098\pm.014$ and an SSIM of $.775\pm.048$. For LPIPS, the render-to-real distance lies between both the blur and noise reference values. For SSIM, render-to-real similarity lies between the two noise levels, but is lower than both blur levels. \notesup{One reason for this could be that LPIPS measures similarity in a learned deep feature space, while SSIM compares luminance, contrast, and local image structure. Illumination differences and slight scene changes between reconstruction and execution may therefore contribute more strongly to the lower SSIM without producing equally large changes in LPIPS.} Qualitative comparisons are provided in Appendix~\ref{app:render_real}.

\section{Limitations}
\label{sec:discussion}
We acknowledge several important limitations. First, WARPED primarily supports rigid-object motion, with an additional extension to a stationary-base articulated object whose manipulated part moves about a single dominant axis of rotation. Extending to more complex articulated-object motion could leverage recent works that distill DINOv2 features~\cite{rsrd, yu2025real2render2realscalingrobotdata}, while deformable objects may be addressed by integrating deformable Gaussian representations~\cite{deformgs}. Next, the method assumes a quasi-static scene in which only the manipulated object moves. Adapting to moving scenes would likely require integrating dynamic Gaussian Splatting representations~\cite{4dgs0, 4dgs1}. \notesup{Our experiments are also conducted in relatively clean tabletop environments, and we do not evaluate the full pipeline under substantial scene clutter.} Additionally, the pipeline assumes relatively short-horizon interactions without repeated re-grasping. Future work could explore explicit contact-state segmentation for handling multi-phase manipulation. Finally, the pipeline fails when the manipulated object becomes fully occluded.

\section{Conclusion}
In this paper, we present WARPED, a framework for learning wrist-camera visuomotor policies from egocentric human demonstration videos by synthesizing wrist-view robot observations and retargeted end-effector trajectories. By combining hand-object tracking, scene reconstruction, and photorealistic rendering, WARPED enables policy training from monocular RGB demonstrations without multiview setups, depth sensors, or custom hardware. Across \notesup{four of five} rigid-object tasks and one articulated-object task, WARPED achieves \notesup{success rates comparable to teleoperation} while reducing demonstration collection time by 
\notesup{4-9 times,}
and improves over teleoperation on the evaluated novel-object tasks. Future work will focus on extending WARPED to more complex articulated and deformable objects, dynamic scenes, and longer-horizon manipulation tasks.


\clearpage
\acknowledgments{This research was supported by the NSF/USDA NIFA AIIRA AI Research Institute under Award 2021-67021-035329.}


\bibliography{references}  

\clearpage
\appendix
\begin{center}
{\Large \bfseries Appendix}
\end{center}
\vspace{1em}

\section{Data Collection}
\subsection{Fiducial Marker}
\label{app_sub:fiducial}
Similar to Universal Manipulation Interface (UMI)~\cite{umi}, a fiducial marker of known size is optionally placed in the scene during the initial scan~(Sec.~\ref{subsec:data_collection}). If present, the marker is used to aid the scene-level scale alignment discussed in Sec.~\ref{subsubsec:interactive_scene_recon}, and is further detailed in Appendix~\ref{app:ssa}. Note that the fiducial marker does not contribute to the Structure-from-Motion (SfM) and is not otherwise used in the rendering pipeline.

\subsection{\label{app_sub:calibration}
Camera Calibration}
The head-mounted egocentric camera used to record the demonstrations is modeled as a pinhole camera and may optionally be calibrated prior to data collection. If calibrated intrinsics are available, they are used directly during localization. Otherwise, the intrinsics are first approximated and then refined during localization by additionally registering each demonstration frame into the existing SfM model and performing bundle adjustment.

\section{Scene-Level Scale Alignment}
\label{app:ssa}
Because SfM reconstructions are ambiguous up to a global scale, the reconstruction from Sec.~\ref{subsec:data_collection} needs to be aligned with the predicted monocular depth maps from Sec.~\ref{subsubsec:interactive_scene_recon}. We estimate the affine mapping between the predicted depth $z^{\text{pred}}$ and SfM depth $z^{\text{sfm}}$
\begin{equation}
    z^{\text{sfm}}\approx A + Bz^{\text{pred}}
\end{equation}
and use the resulting scale and offset to align the scene reconstruction and Gaussian Splat with the monocular depth maps.

For each demonstration frame, we use the 2D-3D SfM correspondences from Hierarchical Localization~\cite{hloc}. Each correspondence gives a 2D keypoint in the image and a matched 3D SfM point, whose camera-frame depth is $z_i^{\text{sfm}}$. The $n$ correspondences whose 2D keypoints are closest to but lie outside the object mask are selected and used to sample the corresponding monocular depth values $z_i^\text{pred}$, where $n$ varies by task but is typically between 10-50. $A$ and $B$ are then estimated by solving the nonlinear least squares equation
\begin{equation}
    \min_{A,B}\; \frac{1}{2}\sum_i\rho\!\left(\left(A + B\,z_i^{\text{pred}} - z_i^{\text{sfm}}\right)^2\right)
\end{equation}
where $\rho$ is the Huber loss function.
If the fiducial marker from Appendix~\ref{app_sub:fiducial} is present, the reconstruction is further rescaled so that the recovered marker geometry matches its known size. This is done by triangulating detected tag corner points across views and scaling the reconstruction such that the reconstructed tag length aligns with its known dimensions.

\section{Hand Pose Initialization}
\label{app:hpi}
The 3D hand pose estimates~(Sec.~\ref{subsubsec:hand_pose_est}) produced by HAMER~\cite{hamer} are often noisy and temporally inconsistent, as HAMER does not use temporal information between frames. To refine and temporally smooth the hand poses, we utilize a two-stage optimization approach. In the first stage, we fix the hand pose parameters $\boldsymbol{\theta}$ and optimize
\begin{equation}
\begin{aligned}
    \mathbf{\Theta}^{\text{coarse}} = \{\mathbf{R}^{hand}, \mathbf{t}^{hand}, \boldsymbol{\beta}\}\\
  \min_{\mathbf{\Theta}^{\text{coarse}}} \lambda_{\text{2D}}\mathcal{L}_{\text{2D}} + \lambda_{\text{smooth}}\mathcal{L}_{\text{smooth}}
\;
\end{aligned}
\end{equation}
where
\begin{equation}
\label{eq:2d_smooth}
\begin{aligned}
    V^{hand} &= \mathbf{R}^{hand}\cdot\text{MANO}(\boldsymbol{\theta}, \boldsymbol{\beta}) + \mathbf{t}^{hand}\\
    \mathcal{L}_{\text{2D}} &= \|\Pi(V^{hand}) - \mathbf{v}_{2d}^{hand}\|\\
    \mathcal{L}_{\text{smooth}} &= \sum_{t}\sum_{\mathbf{v}^h\in V^{hand}}\|\mathbf{v}^h_t - \mathbf{v}^h_{t-1}\|_2^2
\end{aligned}
\vspace{2pt}
\end{equation}
Here, $\Pi(\cdot)$ denotes camera projection, and $\mathbf{v}_{2D}^{hand}$ are the 2D hand keypoints predicted by HAMER. 

In the second stage, we additionally optimize the hand pose parameters $\boldsymbol{\theta}$ and add a depth supervision loss to align the hand vertices with the predicted monocular depth maps
\begin{equation}
\begin{aligned}
    \mathbf{\Theta}^{\text{fine}} &= \{\mathbf{R}^{hand}, \mathbf{t}^{hand}, \boldsymbol{\theta},\boldsymbol{\beta}\}\\
    \mathcal{L}_{\mathcal{D}_{hand}} &= \|(\mathcal{D}^{hand} - \hat{\mathcal{D}}^{hand}) \odot(1 - \hat{\mathcal{M}}^{obj})\|\\
  \min_{\mathbf{\Theta}^{\text{fine}}} \lambda_{\text{2D}}\mathcal{L}_{\text{2D}} + &\lambda_{\text{smooth}}\mathcal{L}_{\text{smooth}} + \lambda_{\mathcal{D}_{hand}}\mathcal{L}_{\mathcal{D}_{hand}}
\;
\end{aligned}
\end{equation}
where $\mathcal{L}_{\text{2D}}$ and $\mathcal{L}_{\text{smooth}}$ are the same as Eq.~\ref{eq:2d_smooth}. For the depth loss, hand vertices are filtered using a depth discontinuity filter and statistical outlier removal. We find this two-stage approach to be more stable than initially optimizing all hand pose parameters with depth supervision. Frames where the hand is outside the camera view are handled by interpolating neighboring valid hand poses. This is sufficient for our setting, as the hand remains visible for most of the demonstration.

\section{Object Pose Initialization}
\label{app:obj_pose_init}
As discussed in Sec.~\ref{subsubsec:object_pose_init}, MegaPose~\cite{megapose} provides an initial 6D pose estimate $(\mathbf{R}_0^{obj*}, \mathbf{t}_0^{obj*})$. This estimate is obtained using the canonical object mesh reconstructed by SAM3D~\cite{sam3d_obj}, with vertices $V^{obj*}$ which are defined up to an arbitrary scale. \notesup{We initialize the object scale using RaySt3R~\cite{rayst3r}; given the RGB image, predicted depth, and object mask, we obtain a RaySt3R point cloud and scale the canonical object mesh so that its PCA extent matches the estimated geometry.} To refine the object mesh, an initial contact frame $\tilde{t}_s$ is first estimated by thresholding the overlap between the hand and object segmentation masks. Then, we optimize
\begin{equation}
    \min_{\mathbf{R}^{obj}_0, \mathbf{t}^{obj}_0, s^{obj}} \; \lambda_{\mathcal{M}_{obj}} \mathcal{L}_{\mathcal{M}_{obj}}
+
\lambda_{\mathcal{D}_{obj}} \mathcal{L}_{\mathcal{D}_{obj}}
\end{equation}
where $\mathbf{R}_0^{obj}$ and $\mathbf{t}_0^{obj}$ are initialized as $\mathbf{R}_0^{obj*}$ and $\mathbf{t}_0^{obj*}$ respectively, and $s^{obj}$ is initialized \notesup{using the RaySt3R scale estimate}. The losses $\mathcal{L}_{\mathcal{M}_{obj}}$ and $\mathcal{L}_{\mathcal{D}_{obj}}$ follow the same definitions as Eq.~\ref{eq:mask_obj_only} and Eq.~\ref{eq:obj_depth}, but are evaluated using the scaled object mesh vertices $V^{obj} = s^{obj}V^{obj*}$ for all frames $t\leq \tilde{t}_s$. During this interval, the object is assumed to be stationary. The estimate $\tilde{t}_s$ is only used for initial object scale and pose refinement. The final contact start and end frames $(t_s, t_e)$ are computed using motion changes~(Appendix~\ref{app:start_end}).

\section{Hand-Object Optimization}

\subsection{Object Pose Estimation}
\label{app_subsec:ope}
Given the object mesh with vertices $V^{obj}$ and faces $F^{obj}$, object masks and monocular depth maps $\hat{\mathcal{M}}^{obj}$, $\hat{\mathcal{D}}^{obj}$, hand masks $\hat{\mathcal{M}}^{hand}$, and the initialized pose $(\mathbf{R}^{obj}_0, \mathbf{t}^{obj}_0)$ from Appendix~\ref{app:obj_pose_init}, a differentiable rasterizer $\mathcal{R}$~\cite{nvdiffrasst} is used to render the image, mask, and depth of the object 
\begin{equation}
\label{eq:obj_rend}
    \mathcal{I}_t^{obj},
    \mathcal{M}_t^{obj}, \mathcal{D}_t^{obj} = \mathcal{R}(\mathbf{R}^{obj}_tV^{obj} + \mathbf{t}^{obj}_t, F^{obj})
\end{equation}

An occlusion-aware mask loss~\cite{phosa} is first calculated between the rendered and SAM2 masks.
\begin{equation}
\label{eq:mask_obj_only}
    \mathcal{L}_{\mathcal{M}_{obj}} = \|(\mathcal{M}_t^{obj} - \hat{\mathcal{M}}_t^{obj}) \odot(1 - \hat{\mathcal{M}}_t^{hand})\|
\end{equation}

Because similar object masks can be produced by different poses, particularly when faced with hand occlusions, we additionally utilize a depth consistency loss that encourages consistency between the rendered and predicted depths.
\begin{equation}
\label{eq:obj_depth}
    \mathcal{L}_{\mathcal{D}_{obj}} = \|(\mathcal{D}_t^{obj} - \hat{\mathcal{D}}_t^{obj}) \odot(1 - \hat{\mathcal{M}}_t^{hand})\|_2^2
\end{equation}

To account for sparsity and noise in monocular depth predictions, we additionally incorporate image-based feature supervision. We extract DINOv2~\cite{dinov2} features from the rendered image $\mathcal{F}_t$ and the masked original frame $\hat{\mathcal{F}}_t$ and compute a cosine similarity loss.
\begin{equation}
    \begin{aligned}
        \mathcal{F}_t &= \mathcal{G}(\mathcal{I}^{obj}_t), \\
        \hat{\mathcal{F}}_t &= \mathcal{G}(\hat{\mathcal{I}}_t \odot \mathcal{M}_t^{obj}), \\
        \mathcal{L}_{\text{DINO}} &=
        \frac{1}{P}
        \sum_{p=1}^{P}
        \left(
        1 -
        \frac{
        \mathcal{F}_{t,p} \cdot \hat{\mathcal{F}}_{t,p}
        }{
        \lVert \mathcal{F}_{t,p} \rVert_2 \,
        \lVert \hat{\mathcal{F}}_{t,p} \rVert_2
        }
        \right)
    \end{aligned}
\end{equation}

\noindent where $\hat{\mathcal{I}}_t$ is the original image, and $\mathcal{G}$ denotes the DINOv2 network with a ViT-S~\cite{vit} backbone, using features extracted from the ninth layer.  

\notesup{For the Wipe Brush and Can on Plate tasks, where the manipulated objects are smaller and more easily occluded, we additionally use CoTracker~\cite{karaev23cotracker, karaev24cotracker3} point tracks to provide image-space supervision. Given $N$ tracked 2D object points $\mathbf{v}^{obj}_{2d,t,i}$ and their corresponding 3D mesh points $\mathbf{v}^{obj}_i \in V^{obj}$, we define}
\begin{equation}
\mathcal{L}_{track}
=
\frac{1}{N}
\sum_{i=1}^{N}
\left\|
\Pi\left(
\mathbf{R}^{obj}_t \mathbf{v}^{obj}_i + \mathbf{t}^{obj}_t
\right)
-
\mathbf{v}^{obj}_{2d,t,i}
\right\|
\end{equation}
\notesup{For these tasks, we add $\lambda_{track}\mathcal{L}_{track}$ in the object pose optimization objective in Eq.~\ref{eq:obj_pose_opt}.}

\subsection{Contact Start and End Estimation}
\label{app:start_end}
Once object poses for all frames have been estimated, we estimate the contact start and end frames $(t_s, t_e)$. Inspired by Lin~\textit{et~al}.~\cite{reisom}, we set a translation threshold $\epsilon_o$ and rotation threshold $\epsilon_r$. If at time $t_s$ the change in object translation $\Delta o$ or rotation $\Delta r$ exceeds these thresholds for $m$ consecutive frames, $t_s$ is determined to be the contact start frame. 
\begin{equation}
    \begin{aligned}
        t_s^o &= \inf \left\{\, t \;:\;
        \min_{k \in \{t,\ldots,t+m-1\}} \left\| \Delta o_k \right\|_2 \ge \epsilon_o
        \right\}\\
        t_s^r &= \inf \left\{\, t \;:\;
        \min_{k \in \{t,\ldots,t+m-1\}} \left\| \Delta r_k \right\|_2 \ge \epsilon_r
        \right\}\\
        t_s &= \text{min}(t_s^o, t_s^r)
    \end{aligned}
\end{equation}
Similarly, the end contact frame $t_e$ is determined by identifying the earliest time at which the object translation \notesup{and} rotation fall below the respective thresholds for 
$m$ consecutive frames.
\begin{equation}
    \begin{aligned}
        t_e^o &= \inf \left\{\, t \;:\;
        \max_{k \in \{t-m+1,\ldots,t\}} \left\| \Delta o_k \right\|_2 < \epsilon_o
        \right\}\\
        t_e^r &= \inf \left\{\, t \;:\;
        \max_{k \in \{t-m+1,\ldots,t\}} \left\| \Delta r_k \right\|_2 < \epsilon_r
        \right\}\\
        t_e &= \max(t_e^o, t_e^r)
    \end{aligned}
\end{equation}

\subsection{Joint Hand-Object Refinement}
\label{app:joint_ho}

To jointly refine the hand and object poses, all parameters are optimized simultaneously across all frames. We denote the object parameters as $\boldsymbol{\Theta}^{obj}$, which consists of the object rotations $\mathbf{R}^{obj}$ and translations $\mathbf{t}^{obj}$ for all frames in addition to a global scaling factor $s^{obj}$. The object is assumed to remain stationary for all $t \leq t_s$ and $t \geq t_e$.
In addition, we optimize the MANO~\cite{mano} hand parameters $\boldsymbol{\Theta}^{hand}$, which include the global hand rotations $\mathbf{R}^{hand}$, global hand translations $\mathbf{t}^{hand}$, hand pose parameters $\boldsymbol{\theta}$, and shared hand shape parameters $\boldsymbol{\beta}$. Our joint hand-object optimization uses the following loss functions to enforce visual and interactive constraints:

\textbf{Occlusion-Aware Mask Loss.}
Similar to Eq.~\ref{eq:obj_rend}, we use a differentiable rasterizer to render masks and depths for both the object and the hand across all frames.
\begin{equation}
    \begin{aligned}
        M^{obj}, D^{obj} &= \mathcal{R}(\mathbf{R}^{obj}V^{obj} + \mathbf{t}^{obj}, F^{obj})\\
        M^{hand}, D^{hand} &= \mathcal{R}(\mathbf{R}^{hand}V^{hand} + \mathbf{t}^{hand}, F^{hand})
    \end{aligned}
\end{equation}
The hand mesh is obtained from the MANO hand model using the current hand parameters.
\begin{equation}
    V^{hand}, F^{hand} = \text{MANO}( \boldsymbol{\theta},\boldsymbol{\beta})
\end{equation}
We compute occlusion-aware mask losses that account for mutual hand-object occlusions.
\begin{equation}
\label{eq:joint_mask}
    \begin{aligned}
        \mathcal{L}_{\mathcal{M}_{obj}} &= \|(\mathcal{M}^{obj} - \hat{\mathcal{M}}^{obj}) \odot(1 - \hat{\mathcal{M}}^{hand})\|\\
        \mathcal{L}_{\mathcal{M}_{hand}} &= \|(\mathcal{M}^{hand} - \hat{\mathcal{M}}^{hand}) \odot(1 - \hat{\mathcal{M}}^{obj})\|\\
        \mathcal{L}_{\mathcal{M}} &= \mathcal{L}_{\mathcal{M}_{obj}} + \mathcal{L}_{\mathcal{M}_{hand}}
    \end{aligned}
\end{equation}

\textbf{Depth Loss.}
We apply the same depth loss in Eq.~\ref{eq:obj_depth} to both the object and the hand for all frames.
\begin{equation}
\label{eq:joint_depth}
    \begin{aligned}
        \mathcal{L}_{\mathcal{D}_{obj}} &= \|(\mathcal{D}^{obj} - \hat{\mathcal{D}}^{obj}) \odot(1 - \hat{\mathcal{M}}^{hand})\|_2^2\\
        \mathcal{L}_{\mathcal{D}_{hand}} &= \|(\mathcal{D}^{hand} - \hat{\mathcal{D}}^{hand}) \odot(1 - \hat{\mathcal{M}}^{obj})\|_2^2\\
        \mathcal{L}_{\mathcal{D}} &= \mathcal{L}_{\mathcal{D}_{obj}} + \mathcal{L}_{\mathcal{D}_{hand}}
    \end{aligned}
\end{equation}

\textbf{Contact Loss.}
Inspired by previous work~\cite{hasson, rhov, hold, reisom}, proximity is encouraged between frequently contacted hand vertices $V^{tip} \subset V^{hand}$ (Fig.~\ref{fig:retarget_and_render}(a)) and object vertices during contact frames to promote interaction. Let
\begin{equation}
\bar{V}^{obj}_t = \mathbf{R}^{obj}_t V^{obj} + \mathbf{t}^{obj}_t,
\qquad
\bar{V}^{hand}_t = \mathbf{R}^{hand}_t V^{hand}_t + \mathbf{t}^{hand}_t,
\qquad
\bar{V}^{tip}_t \subset \bar{V}^{hand}_t
\end{equation}
denote the transformed object and hand vertices at frame $t$. The contact loss is defined as
\begin{equation}
\label{eq:contact}
\mathcal{L}_{\text{contact}}(t) =
\begin{cases}
\displaystyle
\sum_{\bar{\mathbf{v}}^\tau_t \in \bar{V}^{tip}_t}
\min_{\bar{\mathbf{v}}^o_t \in \bar{V}^{obj}_t}
\left\| \bar{\mathbf{v}}^\tau_t - \bar{\mathbf{v}}^o_t \right\|_2^2,
& t_s \leq t \leq t_e, \\[8pt]
0, & \text{otherwise}.
\end{cases}
\end{equation}

\textbf{Collision Loss.}
To discourage physically implausible interpenetration between the hand and the object during contact, we penalize hand vertices that fall inside the object volume. We precompute a truncated signed distance field (TSDF) for the object mesh and evaluate it at each hand vertex
\begin{equation}
\label{eq:tsdf}
\mathcal{L}_{\text{col}}(t) =
\displaystyle
\sum_{\bar{\mathbf{v}}^h_t \in \bar{V}^{hand}_t}
\Phi^{obj}_t(\bar{\mathbf{v}}^h_t)
\end{equation}
where $\Phi^{obj}_t(\cdot)$ denotes the transformed object TSDF at frame $t$ evaluated at hand vertex $\bar{\mathbf{v}}^h_t \in \bar{V}^{hand}_t$.

\textbf{Stable Grasp Loss.}
We use a stable grasp loss~\cite{rhit} that encourages fingertip vertices $\bar{\mathbf{v}}^\tau_t\in \bar{V}^{tip}_t$ to maintain locally consistent distances to object vertices $\bar{\mathbf{v}}^o_t \in \bar{V}^{obj}_t$ during contact
\begin{equation}
\label{eq:stable_grasp}
\mathcal{L}_{\text{sg}} =
\sum_{t=t_s}^{t_e-k}
\sum_{i=1}^{k}
\lVert d^{o\tau}_t - d^{o\tau}_{t+i}\rVert,
\quad
d^{o\tau}_t = \lVert \bar{\mathbf{v}}_t^\tau - \bar{\mathbf{v}}_t^o\rVert_2^2
\end{equation}

The final joint optimization minimizes Eq.~\eqref{eq:joint_ho}.

\subsection{Auxiliary Losses}
\label{app_subsec:aux_loss}

In addition to the losses defined in Appendix~\ref{app:joint_ho}, we include auxiliary losses $\mathcal{L}_{\text{aux}}$ presented in Eq.~\ref{eq:joint_ho} to further guide the hand-object optimization. 

\textbf{Scene TSDF Loss.} 
To prevent collisions with the static scene, a scene-level TSDF is constructed by rendering depth from the scene Gaussian Splat. A collision loss penalizes hand and object vertices that penetrate the scene surface.
\begin{equation}
\label{eq:scene_tsdf}
\mathcal{L}_{\text{scene}} =
\sum_{\mathbf{\bar{v}}^h \in \bar{V}^{hand}} \Phi^{scene}(\mathbf{\bar{v}}^h)
\;+\;
\sum_{\mathbf{\bar{v}}^o \in \bar{V}^{obj}} \Phi^{scene}(\mathbf{\bar{v}}^o),
\end{equation}
Here, $\Phi^{scene}(\cdot)$ denotes the scene TSDF evaluated at hand vertex $\mathbf{\bar{v}}^h \in \bar{V}^{hand}$ and object vertex $\mathbf{\bar{v}}^o \in \bar{V}^{obj}$.

\textbf{Resting-on-Scene Loss.}
When the hand is not in contact with the object, we encourage the object to remain in contact with the static scene. Using $\Phi^{scene}$, we extract a scene point cloud $P^{scene}$ and constrain object mesh vertices to remain close to the scene surface during non-contact frames, reducing floating object artifacts.
\begin{equation}
\mathcal{L}_{\text{rest}}(t) =
\begin{cases}
0, & t_s \le t \le t_e, \\[6pt]
\displaystyle
\sum_{\mathbf{\bar{v}}^o \in \bar{V}^{obj}}
\min_{\mathbf{p} \in P^{scene}}
\left\| \mathbf{\bar{v}}^{o}_{t} - \mathbf{p} \right\|_2^2,
& \text{otherwise.}
\end{cases}
\end{equation}

\textbf{Hand Projection Loss.}
The hand projection loss $\mathcal{L}_{\text{2D}}$ in Eq.~\ref{eq:2d_smooth} constrains the projected hand vertices to remain close to the original 2D keypoints predicted by HAMER.

\textbf{Hand Pose Regularization Loss.}
A hand pose regularization term is included to limit deviations of hand pose $\boldsymbol{\theta}$ from the original HAMER output $\hat{\boldsymbol{\theta}}$. This prevents unnatural hand shapes and maintains physically plausible hand configurations.
\begin{equation}
\mathcal{L}_{\text{hp}} = \lVert \boldsymbol{\theta} - \hat{\boldsymbol{\theta}} \rVert
\end{equation}
The auxiliary loss is defined as the weighted sum of all auxiliary terms described above.
\begin{equation}
\label{eq:aux}
\mathcal{L}_{\text{aux}} =
\lambda_{\text{scene}} \mathcal{L}_{\text{scene}}
+
\lambda_{\text{rest}} \mathcal{L}_{\text{rest}}
+
\lambda_{\text{2D}} \mathcal{L}_{\text{2D}}
+
\lambda_{\text{hp}} \mathcal{L}_{\text{hp}}
\end{equation}

\section{Retargeting and Rendering}

\begin{wrapfigure}[16]{r}{0.46\textwidth}
    \vspace{-5.5\baselineskip}
    \centering
    \includegraphics[width=\linewidth]{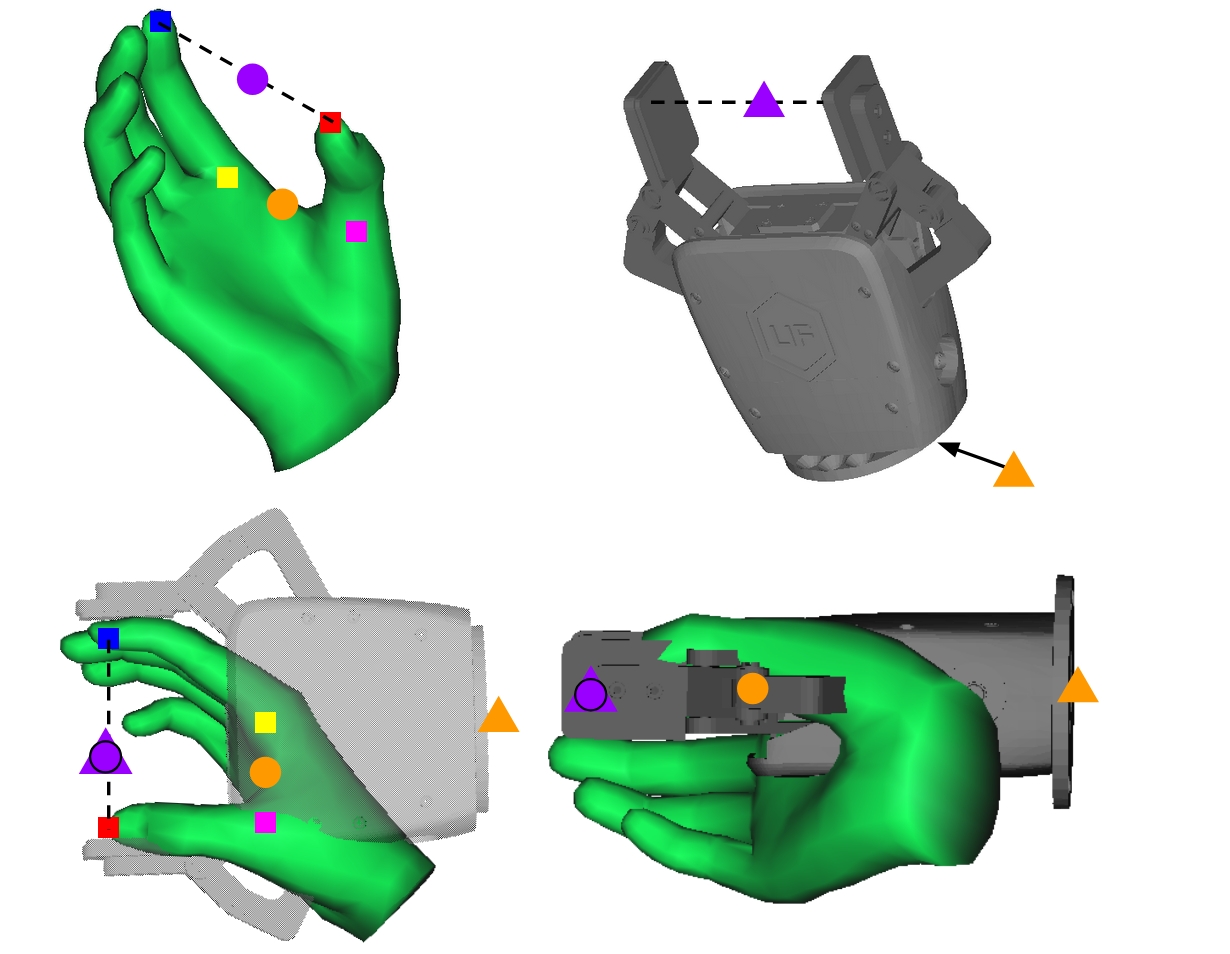}
    
    \setlength{\unitlength}{1cm}
    \begin{picture}(0,0)
        \put(-2.5, 3.0){\footnotesize (a)}
        \put(0.2, 3.0){\footnotesize (b)}
        \put(-2.5, 0.35){\footnotesize (c)}
        \put(0.2, 0.35){\footnotesize (d)}
    \end{picture}
    \vspace{-6pt}
    
    {\small
    {\color{red}$\blacksquare$}~$\text{Th}_{\text{tip}}$ \quad 
    {\color{blue}$\blacksquare$}~$\text{Ind}_{\text{tip}}$ \quad 
    {\color{truemagenta}$\blacksquare$}~$\text{Th}_{\text{mcp}}$ \quad 
    {\color{yellow}$\blacksquare$}~$\text{Ind}_{\text{mcp}}$ \\
    \smallskip
    {\color{realpurple}\scalebox{1.3}{$\bullet$}}~$\text{H}_{\text{tcp}}$ \quad 
    {\color{realpurple}\scalebox{1.2}{$\blacktriangle$}}~$\text{EE}_{\text{tcp}}$ \quad 
    {\color{orange}\scalebox{1.3}{$\bullet$}}~$\text{H}_{\text{base}}$ \quad 
    {\color{orange}\scalebox{1.2}{$\blacktriangle$}}~$\text{EE}_{\text{base}}$
    }

    \caption{Hand-to-end-effector pose mapping. 
    (a) Example hand output showing thumb, index, and derived TCP and base keypoints. (b) Gripper model with corresponding TCP and base keypoints. (c) Front view of the gripper aligned to the hand using the mapped TCP and base keypoints. (d) Side view of the gripper alignment using the TCP and base keypoints.}
    \vspace{-6pt}
    \label{fig:app_map_detail}
\end{wrapfigure}
\subsection{Hand-to-End-Effector Mapping} 
\label{app_subsec: h2eem}

Fig.~\ref{fig:app_map_detail} shows our hand-to-end-effector mapping which is inspired by Papagiannis \textit{et al.}~\cite{papagiannis2025rxretrievalexecutioneveryday}. The tool center point (TCP) of the end-effector $\text{EE}_{\text{tcp}}$ is mapped to the hand as the midpoint $\text{H}_{\text{tcp}}$ between the thumb and index fingertips, $\text{Th}_{\text{tip}}$ and $\text{Ind}_{\text{tip}}$. The vector connecting the thumb and index fingertips defines a principal axis used to orient the gripper such that the gripper base $\text{EE}_{\text{base}}$ is aligned as closely as possible with the midpoint $\text{H}_{\text{base}}$ between the thumb and index MCP joints, $\text{Th}_{\text{mcp}}$ and $\text{Ind}_{\text{mcp}}$.

\subsection{Pre-Contact Trajectory Optimization}
\label{app_subsec_f2:pcto}
As discussed in Sec.~\ref{subsec:retarg_and_red}, pre-contact trajectory optimization is applied to prevent unintended gripper–object collisions. Inspired by Pan \textit{et al.} ~\cite{pan20251001demos}, the optimization combines a funnel loss $\mathcal{L_\text{funnel}}$, collision loss $\mathcal{L_\text{col}}$, and smoothness loss $\mathcal{L_\text{smooth}}$.
$\mathcal{L_\text{funnel}}$ preserves the pre-contact motion by encouraging the trajectories to converge to the same initial end-effector pose.
\begin{equation}
\mathcal{L}_{\text{funnel}}
=
\sum_{t}
w_t
\left\|
\mathbf{t}^{ee}_t - \hat{\mathbf{t}}^{ee}_t
\right\|_2^2,
\end{equation}
Here, $\hat{\mathbf{t}}^{ee}_t$ and $\mathbf{t}^{ee}_t$ denote the original and optimized end-effector translation at timestep $t$ respectively. The weights $w_t$ increase from $w_{\min}$ to $w_{\max}$, which places greater emphasis on matching the demonstration near contact, while allowing more flexibility earlier in the trajectory.
\begin{equation}
w_t
=
w_{\min} + (w_{\max} - w_{\min})
\left(\frac{t}{T-1}\right)^3,
\
\sum_{t=0}^{T-1} w_t = 1
\end{equation}
$\mathcal{L_\text{col}}$ prevents end-effector penetration into the object by penalizing end-effector vertices $V^{ee}$ that lie inside the object surface. This is evaluated using the object TSDF $\Phi^{obj}$.
\begin{equation}
\label{eq:obj_col}
\mathcal{L}_{\text{col}} =
\sum_{\mathbf{v}^{ee} \in V^{ee}} \Phi^{obj}(\mathbf{v}^{ee})
\end{equation}
Lastly, $\mathcal{L_\text{smooth}}$ encourages temporally smooth motion.
\begin{equation}
\label{eq:ee_smooth}
\mathcal{L}_{\text{smooth}}
=
\sum_t
\lVert \mathbf{t}^{ee}_{t+1}-\mathbf{t}^{ee}_t \rVert_2^2
+
\lVert \log\!\big((\mathbf{R}^{ee}_t)^{\top}\mathbf{R}^{ee}_{t+1}\big) \rVert_2^2
\end{equation}
The final optimization is formulated as
\begin{equation}
\min_{{\mathbf{T}^{ee}_{t<t_s}}}
\lambda_{\text{funnel}} \mathcal{L}_{\text{funnel}}
+
\lambda_{\text{col}} \mathcal{L}_{\text{col}}
+
\lambda_{\text{smooth}} \mathcal{L}_{\text{smooth}}
\end{equation}


\subsection{Contact Point Selection}
\label{app_subsec_f3:cps}

Contact points are selected from the object mesh based on Euclidean distance to
the thumb and index fingertips
at the contact start frame $t_s$. For each fingertip, we identify the
50 object mesh vertices closest to the corresponding fingertip vertices~(Fig.~\ref{fig:retarget_and_render}(c-d)), producing
contact sets $V^{\text{contact}}_{\text{thumb}} \subset V^{obj}$ and
$V^{\text{contact}}_{\text{index}} \subset V^{obj}$. The final contact set
$V^{\text{contact}} =
V^{\text{contact}}_{\text{thumb}} \cup
V^{\text{contact}}_{\text{index}}$ is used in
Appendix~\ref{app_subsec_f4:cgr} to optimize the end-effector pose and gripper width so that the
retargeted gripper establishes contact with the object at the corresponding human
contact locations.

\subsection{Contact Grasp Refinement}
\label{app_subsec_f4:cgr}
At the contact start frame $t_s$, the end-effector pose $\mathbf{T}^{ee}_{t_s}$ and gripper width $g_{t_s}$ are refined to ensure a physically plausible grasp~(Fig.~\ref{fig:retarget_and_render}(e-f)).  Using the selected contact points $V^{\text{contact}}$, the end-effector position and gripper width are optimized to align the grasp with the hand-held object
\begin{equation}
\min_{\mathbf{T}^{ee}_{t_s},\, g_{t_s}}
\;
\lambda_{\text{contact}}\mathcal{L}_{\text{contact}}
+
\lambda_{\text{width}}\mathcal{L}_{\text{width}}
+
\lambda_{\text{col}}\mathcal{L}_{\text{col}}
\end{equation}
where
\begin{equation}
\begin{aligned}
\mathcal{L}_{\text{contact}}
&=
\sum_{\mathbf{v}^{ee} \in V^{ee}}
\min_{\mathbf{v}^{c}\in V^{\text{contact}}}
\left\|\mathbf{v}^{ee} - \mathbf{v}^c\right\|_2^2\\
\mathcal{L}_{\text{width}} &= g_{t_s}
\end{aligned}
\end{equation}
and $\mathcal{L}_{\text{col}}$ is the same gripper-object collision loss defined in Eq.~\ref{eq:obj_col}. $\mathcal{L}_\text{width}$ encourages the gripper to be as closed as possible while $\mathcal{L}_{\text{col}}$ prevents gripper-object penetration.

\section{Demonstration Augmentation}
\label{app:augmentation}

As discussed in Sec.~\ref{policy_training}, we apply the following augmentations to increase dataset diversity and improve robustness to out-of-distribution trajectories observed during inference.

\textbf{Object mesh retexturing.}
For each manipulated object, we use GPT-5~\cite{singh2026openaigpt5card} to generate 10 texture augmentation descriptions. These descriptions are passed to TRELLIS~\cite{trellis} to retexture the object meshes. The retextured meshes are used to render wrist-view observations. 

\textbf{Object pose translation.}
The support plane of the object is estimated using the predicted object pose~(Sec.~\ref{subsubsec:object_pose_init}) and the scene point cloud from the initial reconstruction~(Sec.~\ref{subsec:data_collection}). We sample a random translation within $\pm 5$ cm per axis on this plane, which is then applied to the initial object pose and corresponding end-effector trajectory. For tasks with a fixed final object pose, such as Can on Plate, the offset is linearly interpolated back to zero over the contact interval.

\textbf{Initial gripper pose randomization.}
Small rotational and translational perturbations are applied to the initial end-effector pose before contact. Rotations are sampled within $[-3^\circ, 3^\circ]$ about the gripper $x$-axis, $[-5^\circ, 5^\circ]$ about the $y$-axis, and $[-10^\circ, 10^\circ]$ about the $z$-axis, while translations are sampled within $\pm 2$ cm per axis. The perturbed trajectory is interpolated back to the original trajectory over the pre-contact interval.

\textbf{Scale augmentation.}
Random scale perturbations are used to resize both the reconstructed scene and rendered gripper assets before rendering. Scene and gripper scale factors are sampled uniformly from $[0.95, 1.05]$ and $[0.98, 1.02]$, respectively.

\textbf{Wrist-camera perturbation.}
We apply random perturbations to the wrist-camera extrinsics and intrinsics before rendering. Extrinsic perturbations consist of a random rotation within $[-4^\circ, 4^\circ]$ and a translation up to 4 mm. Intrinsic perturbations are applied independently to the fisheye focal lengths, principal point, and distortion coefficients using zero-mean Gaussian noise with standard deviation equal to 2\% of each parameter value.

\section{Training and Evaluation Details}
\label{app:train_eval}

\begin{table*}[t]
\captionsetup{font=small}
\caption{Training and evaluation hyperparameters per task. Img-H: image observation horizon; P-H: proprioception observation horizon; Act-H: action horizon; Freq: rollout frequency (Hz); Speed: rollout frequency relative to data collection; V-Arch: vision encoder architecture; Compute: GPU configuration used for training.}
\centering
\setlength{\tabcolsep}{6pt}
\renewcommand{\arraystretch}{1.2}
\begin{tabular}{lccccccc}
\hline\hline
\textbf{Task} &
\textbf{Img-H} &
\textbf{P-H} &
\textbf{Act-H} &
\textbf{Freq} &
\textbf{Speed} &
\textbf{V-Arch} &
\textbf{Compute} \\
\hline
Rotate Box        & 2 & 2 & 8  & 10 & 0.33 & ViT-B/16 & 4xV100 \\
Pour Mug          & 2 & 2 & 8  & 10 & 0.33 & ViT-B/16 & 4xV100 \\
Bottle from Rack  & 2 & 2 & 12 & 10 & 0.33 & ViT-B/16 & 4xV100 \\
Wipe Brush        & 2 & 2 & 12 & 5 & 0.17 & ViT-B/16 & 4xV100 \\
Can on Plate      & 2 & 2 & 8  & 10 & 0.33 & ViT-B/16 & 4xV100 \\
Open Microwave    & 2 & 2 & 12 & 5  & 0.17 & ViT-B/16 & 4xV100 \\
Can on Plate OOD  & 1 & 2 & 8  & 5  & 0.17 & ViT-L/14 & 4xL40 \\
\hline
\end{tabular}
\label{tab:task_hyperparams}
\vspace{-2pt}
\end{table*}

\begin{table*}[t]
\captionsetup{font=small}
\caption{Common training and evaluation hyperparameters.}
\centering
\setlength{\tabcolsep}{30pt}
\renewcommand{\arraystretch}{1.2}

\begin{threeparttable}
\begin{tabular}{lc}
\hline\hline
\textbf{Hyperparameter} & \textbf{Value} \\
\hline
Image Resolution & $224\times224$ \\
Pretrained Vision Encoder & CLIP~\cite{radford2021learningtransferablevisualmodels} \\
Denoising Steps & 50 \\
\hline
Optimizer & AdamW \\
Base Learning Rate & $3\times10^{-4}$ \\
Learning Rate Scheduler & Cosine Decay \\
Weight Decay & $1\times10^{-6}$ \\
Momentum & $\beta_1,\beta_2=0.95, 0.999$ \\
Warmup Steps & 2000 \\
Training Epochs & 120 \\
Batch Size per GPU & 64 \\
\hline
Proprioception Input & relative eef xyz, relative 6d rotation, binary gripper open/close \\
Action Output & relative eef xyz, relative 6d rotation, binary gripper open/close \\
\hline
\vspace{5pt}
\end{tabular}

\end{threeparttable}

\label{tab:common_hyperparams}
\vspace{-8pt}
\end{table*}

We train a diffusion policy based on the UMI implementation. Task-specific hyperparameters are reported in Table~\ref{tab:task_hyperparams}, while hyperparameters shared across all tasks are listed in Table~\ref{tab:common_hyperparams}. Each demonstration is captured at 30Hz.

At the start of each evaluation trial,
the robot is initialized in a task-dependent configuration such
that the manipulated object is within the wrist-camera field of
view and in reasonable proximity to the end-effector, matching
the human hand conditions observed during demonstration.

\section{Hand-Object Optimization Ablations}
\label{app:ho_ablations}
We ablate losses in the joint hand-object refinement and compare against independently optimized hand and object poses using the object tracking metrics from Table~\ref{table:object_tracking_accuracy}. The experiments include:
\begin{enumerate}
    \renewcommand{\labelenumi}{(\alph{enumi})}

    \item \textbf{Independent Opt. Poses:} Hand and object poses are optimized independently without the joint hand-object refinement from Sec.~\ref{subsec:ho_opt}.

   \item \textbf{No Depth Loss:} The depth loss $\mathcal{L}_{D}$ from Eq.~\ref{eq:joint_depth} is removed.

    \item \textbf{No Contact and Collision Loss:} The contact and collision losses $\mathcal{L}_{contact}$ and $\mathcal{L}_{col}$ from Eqs.~\ref{eq:contact} and~\ref{eq:tsdf} are removed.
    
    \item \textbf{No Stable Grasp Loss:} The stable grasp loss $\mathcal{L}_{sg}$ from Eq.~\ref{eq:stable_grasp} is removed.
    
    \item \textbf{No Auxiliary Loss:} The auxiliary loss $\mathcal{L}_{aux}$ from Eq.~\ref{eq:aux} is removed.

    \item \textbf{Full (Ours):} Uses the complete joint hand-object refinement described in Sec.~\ref{subsec:ho_opt}.
\end{enumerate}
RPE is evaluated over a 0.1s horizon. As shown in Table~\ref{table:ho_ablations}, the full objective achieves the lowest ATE, translational RPE, and MPE, although several ablations have overlapping standard deviations. 

\begin{table*}[t]
\caption{Joint hand-object optimization ablations. ATE, RPE Trans, and MPE are in cm; RPE Rot is in degrees.}
\centering
\setlength{\tabcolsep}{9pt}
\renewcommand{\arraystretch}{1.12}
\resizebox{\textwidth}{!}{%
\begin{tabular}{@{}lcccc@{}}
\hline\hline
\textbf{Method} &
\textbf{ATE $\downarrow$} &
\textbf{RPE Trans. $\downarrow$} &
\textbf{RPE Rot. $\downarrow$} &
\textbf{MPE $\downarrow$} \\
\hline
Independent Opt. Poses
& $2.80 \pm 1.13$
& $1.19 \pm 0.43$
& $6.08 \pm 3.19$
& $4.21 \pm 3.06$ \\
No Depth Loss
& $1.63 \pm 0.87$
& $0.74 \pm 0.35$
& $5.10 \pm 3.04$
& $2.57 \pm 3.15$ \\
No Contact and Collision Loss
& $1.47 \pm 0.90$
& $0.73 \pm 0.35$
& $5.20 \pm 3.03$
& $2.31 \pm 3.38$ \\
No Stable Grasp Loss
& $2.14 \pm 2.21$
& $1.00 \pm 0.43$
& $7.35 \pm 2.63$
& $2.60 \pm 3.33$ \\
No Auxiliary Loss
& $1.67 \pm 0.62$
& $0.74 \pm 0.34$
& $\mathbf{4.96 \pm 3.28}$
& $1.86 \pm 0.76$ \\
Full (Ours)
& $\mathbf{1.40 \pm 0.55}$
& $\mathbf{0.70 \pm 0.35}$
& $5.10 \pm 3.12$
& $\mathbf{1.76 \pm 0.96}$ \\
\hline
\end{tabular}%
}
\vspace{-5pt}
\label{table:ho_ablations}
\end{table*}

\section{Object Pose Tracking Visualization}
\label{app:obj_pose_track_vis}

We provide selected qualitative visualizations in Fig.~\ref{fig:fp_comp} comparing object pose tracking from FoundationPose~\cite{wen2024foundationposeunified6dpose} and the proposed hand-object optimization from Sec.~\ref{subsec:ho_opt}. The selected examples show cases where FoundationPose is unable to accurately track the object. 

\section{Render-to-Real Fidelity Visualization}
\label{app:render_real}
\begin{figure*}[h]
    \centering
    \includegraphics[width=0.97\textwidth]{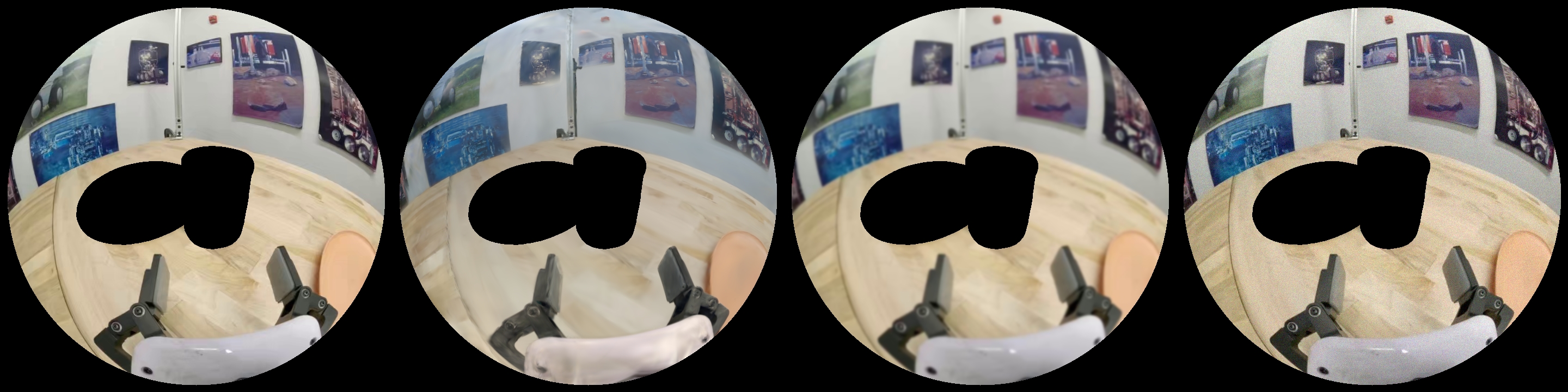}
    \setlength{\unitlength}{1cm}
    \begin{picture}(0,0)
        \put(-13.55, 3.0){\footnotesize\textcolor{white}{(a)}}
        \put(-10.26, 3.0){\footnotesize\textcolor{white}{(b)}}
        \put(-6.87, 3.0){\footnotesize\textcolor{white}{(c)}}
        \put(-3.48, 3.0){\footnotesize\textcolor{white}{(d)}}
    \end{picture}
    \caption{Median-LPIPS Can on Plate example: (a) real wrist-camera image, (b) rendered wrist-camera view, (c) Gaussian blur ($\sigma_b=2$), and (d) Gaussian noise ($\sigma_n=10/255$).}
    \vspace{-2pt}\label{fig:render_real}
\end{figure*}

We provide a qualitative visualization in Fig.~\ref{fig:render_real} showing the median-LPIPS Can on Plate example, including the real wrist-camera image, the corresponding rendered view, and the Gaussian blur and noise reference perturbations used in Sec.~\ref{subsec:results}.

\section{Additional Experiments}

We run the following additional experiments on the five rigid-object tabletop tasks: Rotate Box, Pour Mug, Bottle from Rack, Wipe Brush, and Can on Plate.

\subsection{Data Augmentation Ablation}
We evaluate the impact of data augmentation from Sec.~\ref{policy_training} by removing augmented demonstrations from training. As shown in Table~\ref{table:app_exp}, data augmentation has a substantial effect on performance. Without augmentation, WARPED fails entirely on three of five tasks, and performance of Can on Plate drops by more than 50\%.
This suggests that the augmented variations are important for bridging the gap between synthetic demonstrations and real robot executions.

\subsection{Robustness to Background Distractors}
\begin{figure}[h]
    \centering
\includegraphics[width=01.0\linewidth]{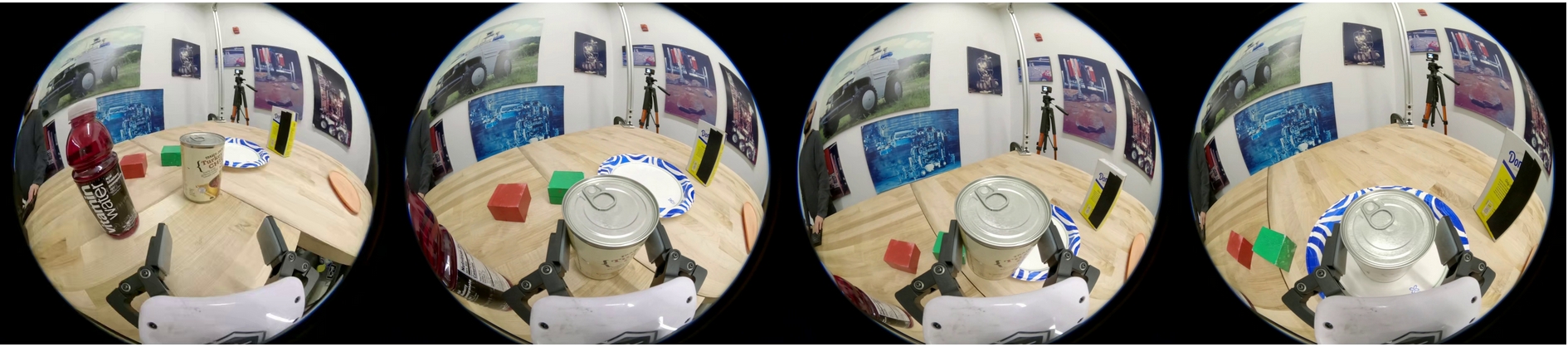}
    \caption{Example of background distractor rollout for Can on Plate task.}
    \label{fig:bg_dist}
    \vspace{-5pt}
\end{figure}

We test the robustness of WARPED when presented with background visual distractors that are not 
present 
in the initial Gaussian Splat of the scene~(Fig.~\ref{fig:bg_dist}). As shown in Table~\ref{table:app_exp}, background distractors have negligible impact on Can on Plate and Bottle from Rack tasks, and result in a small drop in performance for Rotate Box and Wipe Brush. The largest degradation is observed for Pour Mug, where failures are primarily due to missed grasps. 

\begin{table*}[t]
\caption{Success rates for additional experiments.}
\centering
\setlength{\tabcolsep}{9pt}
\renewcommand{\arraystretch}{1.12}
\resizebox{\textwidth}{!}{%
\begin{tabular}{@{}lccccc@{}} 
\hline\hline
\textbf{Method} & \textbf{Rotate Box} & \textbf{Pour Mug} & \textbf{Bottle from Rack} & \textbf{Wipe Brush} & \textbf{Can on Plate} \\ 
\hline
Teleoperation & 16/20 & 19/20 & 16/20 & \textbf{15/20} & 19/20 \\
WARPED (no augmentation) & 0/20 & 17/20 & 0/20 & 0/20 & 8/20 \\
WARPED (background distractors) & 18/20 & 15/20 & \textbf{17/20} & 9/20 & 17/20 \\
WARPED & \textbf{20/20} & 18/20 & \textbf{17/20} & 11/20 & 17/20 \\
Teleoperation + WARPED & 19/20 & \textbf{20/20} & \textbf{17/20} & 11/20 & \textbf{20/20} \\
\hline
\end{tabular}%
}
\vspace{-5pt}
\label{table:app_exp}
\end{table*}

\subsection{Co-training with Teleoperated Demonstrations}
We evaluate whether WARPED demonstrations can be co-trained with teleoperated data. For each task, we combine 15 teleoperated and 15 WARPED demonstrations, including all augmentations. To align human and robot trajectories, human demonstrations are slowed down by a factor of three.
The results are shown in Table~\ref{table:app_exp}. For Rotate Box, Pour Mug, Bottle from Rack, and Can on Plate, co-training achieves performance comparable to or better than training with teleoperation alone. However, for Wipe Brush, performance decreases under co-training. We observe that this task involves more complex motion, and the rendered trajectories from WARPED are more dissimilar from the teleoperated executions, leading to data inconsistencies. 

\subsection{Out-of-Distribution Scenes}
\begin{figure*}[t]
    \centering
    \includegraphics[width=0.97\textwidth]{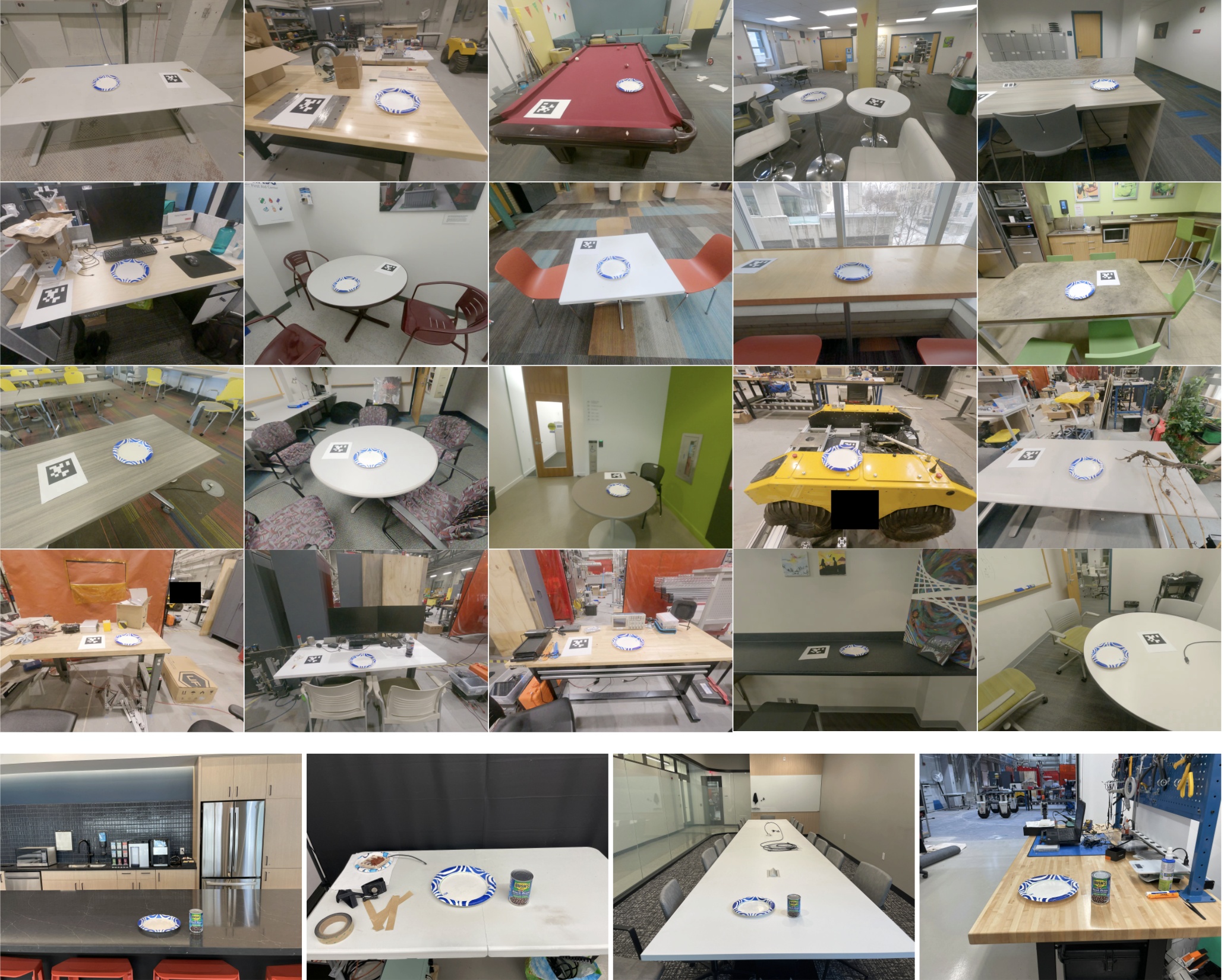}
    \caption{Scenes used in the out-of-distribution experiments. The top four rows show training scenes, and the bottom row shows scenes used for evaluation.}
    \label{fig:ood_vis}

    \vspace{6pt}
    \includegraphics[width=0.97\textwidth]{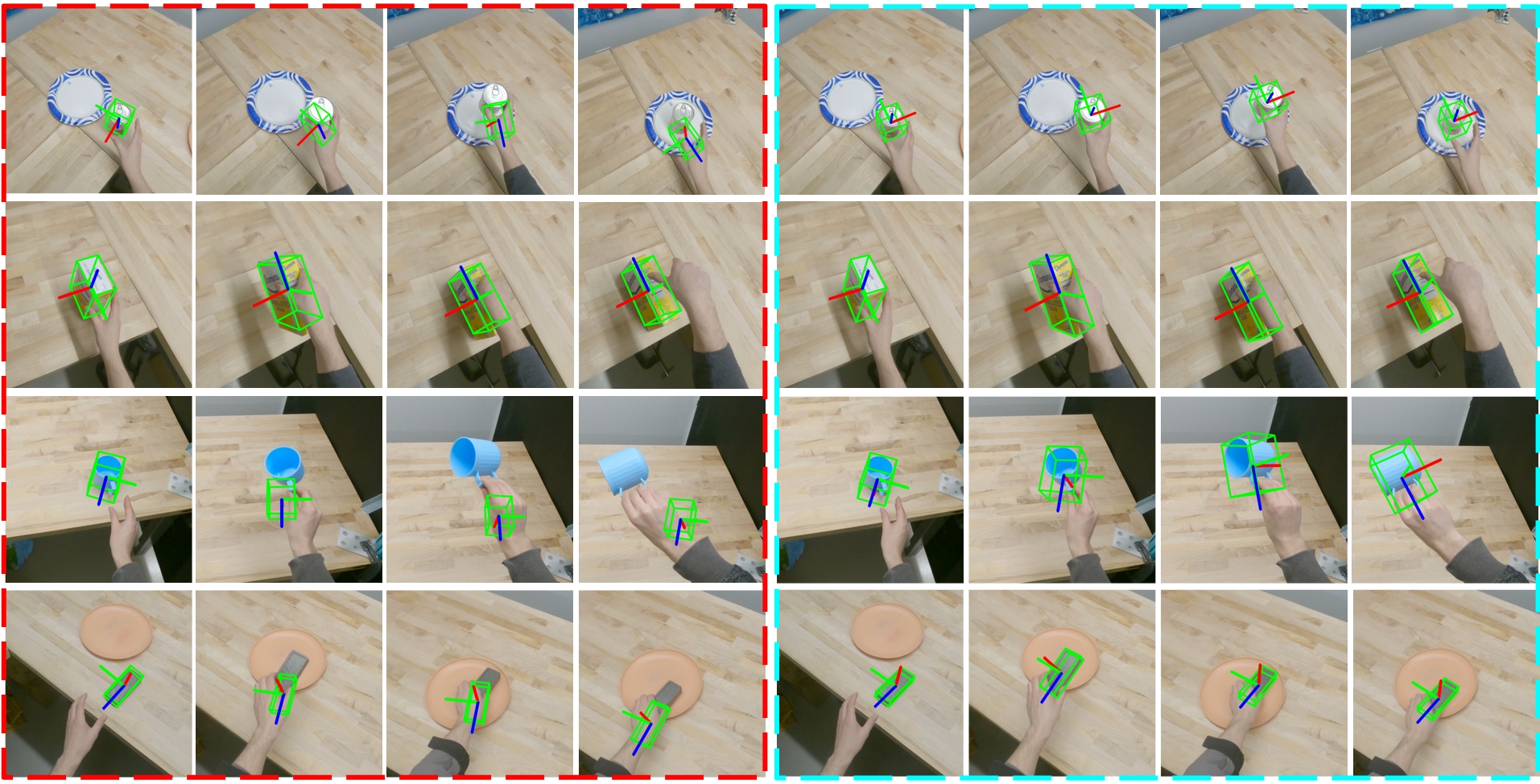}
    
    
    {\small
    {\color{red}\rule[0.5ex]{1.5em}{1.5pt}} \hspace{-1.5em}
    {\color{red}\rule[0.5ex]{0.3em}{1.5pt}\hspace{0.1em}\rule[0.5ex]{0.3em}{1.5pt}\hspace{0.1em}\rule[0.5ex]{0.3em}{1.5pt}} 
    ~FoundationPose \qquad \qquad
    {\color{electriccyan}\rule[0.5ex]{1.5em}{1.5pt}} \hspace{-1.5em}
    {\color{electriccyan}\rule[0.5ex]{0.3em}{1.5pt}\hspace{0.1em}\rule[0.5ex]{0.3em}{1.5pt}\hspace{0.1em}\rule[0.5ex]{0.3em}{1.5pt}}
    ~Hand-Object Optimization
    }


    \captionsetup{justification=raggedright,singlelinecheck=false}
    \caption{Qualitative comparison of object pose tracking between FoundationPose and the hand–object optimization used by WARPED. Hand–object optimization more reliably tracks object pose.
    }
    \label{fig:fp_comp}
\end{figure*}

We examine WARPED's ability to generalize to out-of-distribution scenes. For the Can on Plate task, we collect 50 demonstrations across 20 diverse tabletop settings and evaluate on four unseen scenes for a total of 20 trials. WARPED achieves a success rate of 16/20. 
The variation across scenes is likely due to subtle differences in the training data, including lighting conditions and scene layout. Performance could be 
improved by integrating scene-level augmentation strategies~\cite{yu2025real2render2realscalingrobotdata, yang2025noveldemonstrationgenerationgaussian, pan20251001demos}. Visualizations of training and testing scenes are shown in Fig.~\ref{fig:ood_vis}.

\begin{wraptable}{r}{0.42\linewidth}
\vspace{-32pt}
\centering
\scriptsize
\setlength{\tabcolsep}{6pt}
\renewcommand{\arraystretch}{1.05}
\caption{UMI and WARPED success rates.}
\vspace{-5pt}
\label{table:umi_comparison}
\resizebox{\linewidth}{!}{%
\begin{tabular}{@{}lccc@{}}
\hline\hline
\textbf{Method} & \textbf{Object} &
\shortstack[c]{\rule{0pt}{1.8ex}\textbf{Can on}\\\textbf{Plate}} &
\shortstack[c]{\textbf{Rotate}\\\textbf{Box}} \\
\hline
\multirow{2}{*}{UMI} & Train & \textbf{9/10} & 2/10 \\
                     & Novel & \textbf{10/10} & 0/10 \\
\hline
\multirow{2}{*}{WARPED} & Train & \textbf{9/10} & \textbf{10/10} \\
                        & Novel & 7/10 & \textbf{6/10} \\
\hline
\end{tabular}%
}
\vspace{-5pt}
\end{wraptable}
\subsection{Comparison with UMI}
\label{app:umi_comp}
We include a limited comparison to UMI-style data collection on the Can on Plate and Rotate Box tasks. Policies are trained on 30 demonstrations and evaluated over 20 trials, with 10 using the demonstration object and 10 using a  novel object. Both methods are evaluated on a UR5 with the UMI gripper.

Results are shown in Table~\ref{table:umi_comparison}. For Can on Plate, UMI achieves 9/10 and 10/10 success on the training and novel objects, respectively, while WARPED achieves 9/10 and 7/10. The strong UMI performance is likely because of the similar overhead visual appearance of cylindrical aluminum cans, resulting in consistent and representative demonstrations.

\notesup{For Rotate Box, UMI achieves 2/10 and 0/10 success on the training and novel objects, respectively, while WARPED achieves 10/10 and 6/10. We observe that UMI failures generally reach the object but fail to transition into the subsequent rotational motion, whereas successful WARPED rollouts complete this transition across variations in the relative object and wrist-camera configuration. One possible explanation is that this task is particularly sensitive to deviations in the object, gripper, and wrist-camera configuration near contact, as the policy must execute a large change in end-effector and object orientation. WARPED augments object position, gripper pose, and wrist-camera viewpoint together with the corresponding trajectories, providing greater variation around the collected demonstrations than the unaugmented UMI data.}



\section{Data Processing Times}
\label{app:processing_times}

Data processing times averaged across all tasks and demonstrations are reported in Table~\ref{tab:processing_time}. 
This includes operations performed once per task and once per demonstration. 
Augmentation is reported separately, as it is performed multiple times per demonstration, while retexturing can be reused across demonstrations. 
Reported timings correspond to an unoptimized sequential implementation and depend on the specific CPU and GPU used.

In the current implementation, offline processing requires approximately seven minutes per demonstration, which exceeds the time needed to collect an individual teleoperated demonstration. 
However, this processing reflects automated computation rather than human effort. 
Unlike teleoperation, where collection requires hands-on operator time and robot access, WARPED processing requires no human supervision and can be run offline and in parallel across demonstrations, reducing the active human data collection time~(Table~\ref{table:data_collection_time}). 
Thus, WARPED shifts the primary bottleneck from human-operated robot data collection to automated computation. 
Future work will focus on reducing processing time through parallelization and pipeline-level optimizations.

\FloatBarrier

\begin{table*}[t]
\captionsetup{font=small}
\vspace{-60pt}
\caption{Data processing times averaged across all tasks and demonstrations on an AMD Ryzen 9 7950X CPU and an NVIDIA GeForce RTX 3090 GPU. Timings reflect a sequential, unoptimized implementation.}
\centering
\setlength{\tabcolsep}{20pt}
\renewcommand{\arraystretch}{1.25}

\begin{threeparttable}
\begin{tabular}{llc}
\hline\hline
\textbf{Stage} & \textbf{Operation} & \textbf{Time} \\
\hline

\multicolumn{3}{l}{\textit{Once per task}} \\
\hline
Scene Reconstruction
& Structure-from-Motion & 04:15 \\
& Scene Gaussian Splat (15000 steps) & 04:05 \\
& \textbf{Total (Scene Reconstruction)} & \textbf{08:20} \\
\hline

Object Mesh Reconstruction
& SAM3D Object Mesh Build & 00:17 \\
& Mesh Rendering & 00:24 \\
& Object Gaussian Splat (7000 steps) & 01:07 \\
& \textbf{Total (Object Mesh Reconstruction)} & \textbf{01:48} \\
\hline

\multicolumn{3}{l}{\textit{Per demonstration}} \\
\hline
Interactive Scene Initialization
& Localization & 00:33 \\
& SpatialTrackerV2 Monocular Depth Prediction & 00:20 \\
& Scene-Level Scale Alignment & 00:06 \\
& \textbf{Total (Scene Initialization)} & \textbf{00:59} \\
\hline

Hand Pose Initialization
& Hand Detection & 00:16 \\
& Hand Pose Refinement & 00:29 \\
& \textbf{Total (Hand Pose Initialization)} & \textbf{00:45} \\
\hline

Object Pose Initialization
& Grounding DINO Object Detection & 00:02 \\
& SAM2 Mask Propagation & 00:10 \\
& MegaPose Pose Estimation & 00:08 \\
& Object Pose Refinement & 00:28 \\
& \textbf{Total (Object Pose Initialization)} & \textbf{00:48} \\
\hline

Hand-Object Optimization
& Object Pose Estimation & 01:19 \\
& Joint Hand-Object Refinement & 02:33 \\
& \textbf{Total (Hand-Object Optimization)} & \textbf{03:52} \\
\hline

Retargeting and Rendering
& Pre-Contact Trajectory Optimization & 00:07 \\
& Contact Grasp Refinement & 00:12 \\
& Wrist-Camera Rendering & 00:14 \\
& \textbf{Total (Retargeting and Rendering)} & \textbf{00:33} \\
\hline

& \textbf{Total (Per demonstration)} & \textbf{06:57} \\
\hline

\multicolumn{3}{l}{\textit{Augmentation}} \\
\hline
Shared Across Augmentations
& Retexturing (TRELLIS~\cite{trellis}) & 00:20 \\
& Mesh Rendering & 00:24 \\
& Object Gaussian Splat (7000 steps) & 01:07 \\
\hline
Per-Augmentation Pass
& Pose and Viewpoint Randomization & 00:04 \\
& Wrist-Camera Rendering & 00:14 \\
& \textbf{Total (Per augmentation pass)} & \textbf{00:18} \\
\hline
\end{tabular}

\end{threeparttable}

\label{tab:processing_time}
\end{table*}

\end{document}